\pdfoutput=1
\documentclass[11pt]{wlscirep}
\usepackage[utf8]{inputenc}
\usepackage[T1]{fontenc}
\usepackage{hyperref}
\usepackage{bm}
\usepackage{epstopdf}
\usepackage{makecell}
\usepackage{graphicx}
\usepackage{lscape}
\usepackage{pdflscape}
\usepackage{multirow}
\usepackage{amsmath}
\usepackage{makecell}
\usepackage{filecontents}
\usepackage{times}
\usepackage{epsfig}
\usepackage{amssymb}
\usepackage{threeparttable}  
\usepackage{float}
\usepackage{epsfig}
\usepackage{balance}
\usepackage{booktabs}
\usepackage{enumitem}
\usepackage{mathtools}
\usepackage{makecell}
\usepackage{multirow}
\usepackage{pifont}
\usepackage{stackengine}
\usepackage{xcolor}
\usepackage{colortbl}
\usepackage{tabu}
\usepackage{geometry}
\usepackage{fancyhdr}
\usepackage{array}
\usepackage{subcaption}
\usepackage{caption}
\usepackage{color}
\usepackage{cleveref}   
\usepackage{xspace}
\usepackage[breakable]{tcolorbox}
\usepackage{titletoc}
\usepackage{setspace} 

\usepackage{lineno}
\title{Advancing Real-time Pandemic Forecasting Using Large Language Models: A COVID-19 Case Study}

\newcommand\ours{PandemicLLM\xspace}

\newcommand{\be}{\begin{equation}}
\newcommand{\ee}{\end{equation}}
\newcommand{\ba}{\begin{align}}
\newcommand{\ea}{\end{align}}

\def\ba{{$\bm{a}$}}

\def\model{PandemicLLM\xspace}
\def\models{PandemicLLMs\xspace}
\def\target{HTC\xspace}
\def\metric{WMSE\xspace}

\definecolor{myblue}{RGB}{0, 133, 186}
\definecolor{mygreen}{RGB}{72, 194, 0}
\definecolor{myred}{RGB}{177, 0, 28}
\definecolor{good}{rgb}{0.11, 0.77, 0.11}
\definecolor{bad}{rgb}{0.77, 0.11, 0.11}

\author[1, 2+]{Hongru Du}
\author[3, 4+]{Jianan Zhao}
\author[1, 2+]{Yang Zhao}
\author[1, 2]{Shaochong Xu}
\author[5, 6]{Xihong Lin}
\author[7*]{Yiran Chen}
\author[1, 2, 8*]{Lauren M. Gardner}
\author[1, 2, 7*]{Hao (Frank) Yang}

\affil[1]{Center for Systems Science and Engineering, Johns Hopkins University,
Baltimore, MD, USA.}
\affil[2]{Department of Civil and Systems Engineering, Johns Hopkins University, Baltimore, MD, USA.}
\affil[3]{Mila - Quebec AI Institute, Montréal, QC, Canada.}
\affil[4]{Department of Computer Science, Universite de Montréal, Montréal, QC, Canada.}
\affil[5]{Department of Biostatistics, Harvard T.H. Chan School of Public Health, Boston, MA, USA.}
\affil[6]{ Department of Statistics, Harvard University, Cambridge, MA, USA.}
\affil[7]{Department of Electrical and Computer Engineering, Duke University, Durham, NC, USA.}
\affil[8]{Department of Epidemiology, Johns Hopkins Bloomberg School of Public Health, Baltimore, MD, USA.}

\small{\affil[+]{The authors contributed equally.}
\affil[*]{The corresponding authors information: yiran.chen@duke.edu, l.gardner@jhu.edu, haofrankyang@jhu.edu}}

{ 
\begin{abstract}
\fontsize{11pt}{12pt}\selectfont

Forecasting the short-term spread of an ongoing disease outbreak is a formidable challenge due to the complexity of contributing factors, some of which can be characterized through interlinked, multi-modality variables such as epidemiological time series data, viral biology, population demographics, and the intersection of public policy and human behavior. Existing forecasting model frameworks struggle with the multifaceted nature of relevant data and robust results translation, which hinders their performances and the provision of actionable insights for public health decision-makers. Our work introduces PandemicLLM, a novel framework with multi-modal Large Language Models (LLMs) that reformulates real-time forecasting of disease spread as a text reasoning problem, with the ability to incorporate real-time, complex, non-numerical information—such as textual policies and genomic surveillance data—previously unattainable in traditional forecasting models. This approach, through a unique AI-human cooperative prompt design and time series representation learning, encodes multi-modal data for LLMs. By redefining the forecasting process as an ordinal classification task, PandemicLLM yields more robust and trustworthy predictions, facilitating public health decision-making. The model is applied to the COVID-19 pandemic, and trained to utilize textual public health policies, genomic surveillance, spatial, and epidemiological time series data, and is subsequently tested across all 50 states of the U.S. for a duration of 16 weeks. Empirically, PandemicLLM is shown to be a high-performing pandemic forecasting framework that effectively captures the impact of emerging variants and can provide timely and accurate predictions. The proposed PandemicLLM opens avenues for incorporating various pandemic-related data in heterogeneous formats and exhibits performance benefits over existing models. This study illuminates the potential of adapting LLMs and representation learning to enhance pandemic forecasting, illustrating how AI innovations can strengthen pandemic responses and crisis management in the future.

\end{abstract}
\begin{document}
\flushbottom
\maketitle
\def\graphmodality{detector graph modality}
\def\gridmodality{GPS grid modality}

\newpage

\section{Introduction} \label{sec:intro}

Pandemic forecasting is essential for providing situational awareness and supporting decision-making for policymakers during public health emergencies. The ability to forecast short-term disease outcomes is crucial for informing resource allocation and risk mitigation strategies in near-term time frames, ultimately aiming to minimize the burden of diseases. Predominantly, the existing forecasting models used in research and practice can be categorized into two types: mechanistic models, which simulate transmission dynamics in the population through compartmental models such as the SIR model and its derivatives\cite{chang2021mobility, giordano2020modelling, ihme2021modeling}, and statistical models, which adopt data-driven approaches to forecast disease trends using historical data patterns\cite{gao2020machine, bracher2021pre,li2023wastewater, du2023incorporating}. Different models are suited for distinct forecasting needs: Mechanistic models are beneficial for long-term projections due to their ability to integrate scenario assumptions, whereas statistical models are helpful for short-term forecasting because they can adapt immediate trends effectively \cite{reich2022collaborative}. Despite the pivotal role of statistical models in short-term forecasting,
they are limited in their ability to 1) adapt to multi-modal data in real-time\cite{rosenfeld2021epidemic, sun2023public}, 2) respond to rapidly changing policies \cite{hsiang2020effect}, 3) account for the emergence of new variants \cite{li2021emergence}, and 4) translate the predicted results into useful decision-support guidance with trustworthiness \cite{nixon2022evaluation}.
Consequently, current pandemic forecasting models have often struggled to identify and predict critical turning points in the pandemic's trajectory \cite{castro2020turning, cramer2022evaluation, friedman2021predictive}. Moreover, the absence of a transparent interpretation of these results can diminish public trust in these models and even potentially undermine the efficacy of public health responses.


The COVID-19 pandemic highlighted each of these deficiencies in the existing set of disease forecasting tools, which, as a result, struggled to accurately forecast disease spreading patterns\cite{ioannidis2022forecasting}. In particular, the complexities of the COVID-19 pandemic arose from several interrelated factors that were difficult to incorporate into predictive models: 1) the virological characteristics of the virus\cite{telenti2021after}, such as relative transmissibility, severity, and impact on immunity, 2) the diverse demographic profiles of affected populations, including their evolving immunity \cite{nepomuceno2020besides}, and 3) the dynamic relationship between public policy and human behavior \cite{ruggeri2023synthesis}. The virological dynamics, including factors like the characteristics of viruses and their current prevalence, represent a biological ``language'' with its own syntax and semantics, ruling how a virus behaves and evolves \cite{searls2002language}. Simultaneously, human dynamics, including demographics, behavioral patterns, and responses to policy changes, are akin to a socio-cultural ``language'', rich with subtle cognitive processes. Recognizing these complexities, we aim to enhance disease forecasting by exploiting the richness of timely, complex data made available in multiple structures and formats, diverging from the current research trend of simplifying information into solely numerical forms for the purposes of mathematical modeling.

\begin{figure}[h!]
    \centering
    \includegraphics[width=0.94\textwidth]{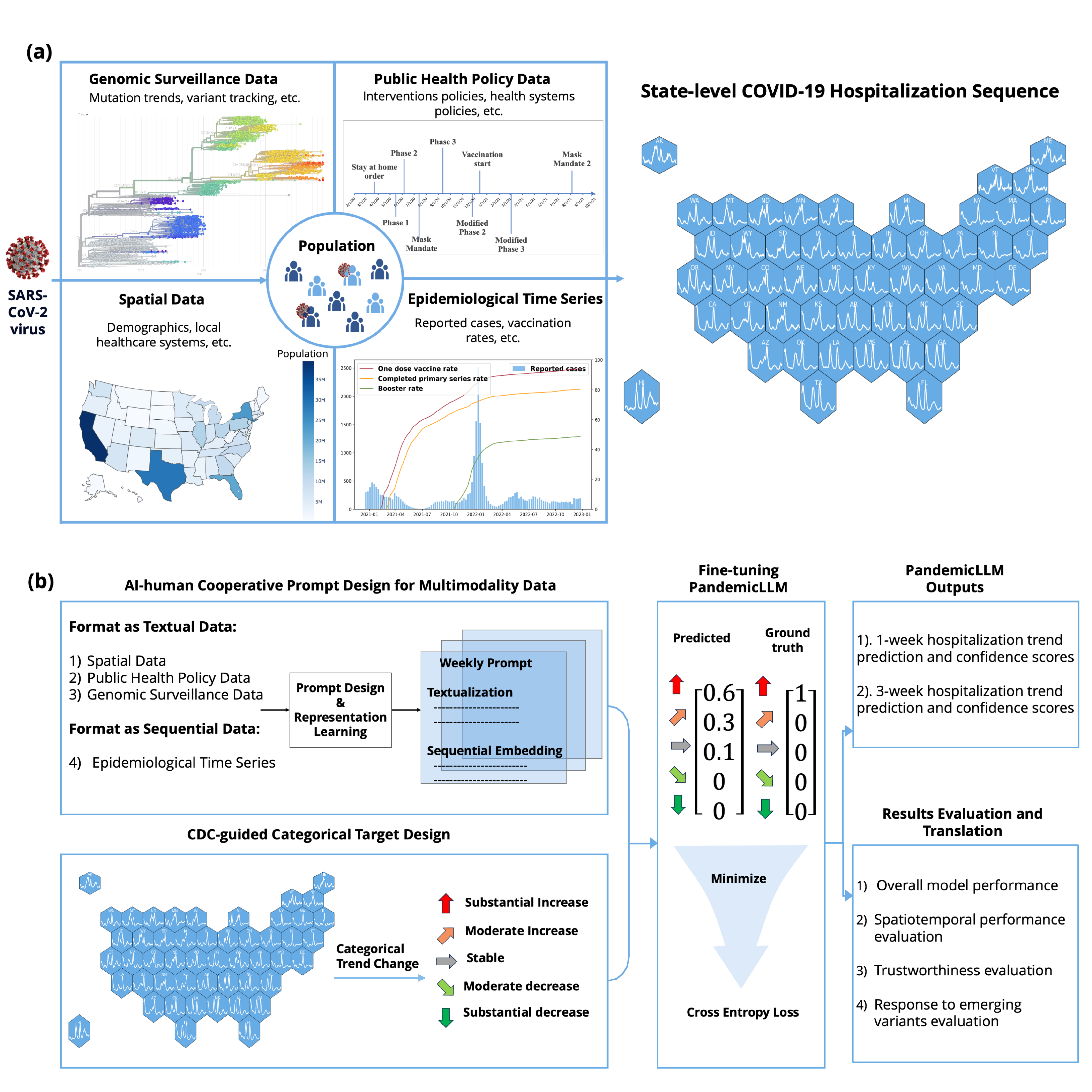}
    \caption{\small{\textbf{The overview of \models' pandemic data streams and pipeline.} \textbf{(a) Multi-modality data insights into Pandemic.} Our multi-modality dataset integrates four types of pandemic data sources: spatial, epidemiological time series, public health policy, and genomic surveillance data. Spatial data includes demographic and healthcare indicators, whereas the epidemiological time series aspect covers reported cases, hospitalizations, and vaccination rates. Data about policy detail governmental interventions in a textual format, and the genomic surveillance data integrates textual descriptions of variants with weekly sequences regarding their prevalence. The data comprises 5,200 records, covering all 50 U.S. states over 104 weeks. The phylogenetic tree of SARS-CoV-2 was generated using Nextstrain\cite{hadfield2018nextstrain}. \textbf{(b) \models' construction pipeline.} To forecast pandemic hospitalization trends, we formulate the problem as an ordinal classification task. We define five categories following CDC guidance\cite{cdc_tracker}: Substantial Decrease, Moderate Decrease, Stable, Moderate Increase, and Substantial Increase. By converting multi-modality data into a text format through AI-human cooperative prompt design, \models are fine-tuned with these prompts and targets for 1-week and 3-week forecasts. We emphasize rigorous performance assessment to verify the accuracy and trustworthiness of our predictions.}}
    \label{fig: fig_1}
\end{figure}

Recently, the Large Language Models (LLMs) stand out as a substantial advancement in artificial intelligence (AI), which have demonstrated proficiency in multi-modal contextual feature learning \cite{singhal2023large, yang2022large, jiang2023health, beaulieu2023predicting, thirunavukarasu2023large}. Their strength in text generation and reasoning suggests potential for understanding the complex dynamics of disease spread \cite{bzdok2024data}.
However, LLMs are designed for processing information in natural language, which presents the following distinct challenges when applying them to pandemic forecasting: 1) Incorporating multi-modality information: The data used for forecasting pandemics include epidemiological time series, public health policy, genomic surveillance, and demographics profiles, requiring these multi-modality inputs to be organized and tailored into suitable learned representations and structured contextual prompts. 2) Incapability in modeling temporal dynamics: LLMs typically struggle with time series data due to the tokenization of continuous numbers \cite{LLM_time}, requiring new methodologies that strategically embed temporal information into prompts, enabling LLMs to effectively utilize these information for reasoning.

In light of the limitations of existing models and the well established text reasoning capability of LLMs, we propose PandemicLLM, the first LLM-based framework for pandemic forecasting. We reformulate pandemic forecasting as a text reasoning task, allowing the integration of new data streams that were not previously utilized in pandemic forecasting models. The proposed framework is designed for state-level COVID-19 hospitalization forecasting across the U.S., targeting prediction horizons of 1-week and 3-week. We tackle the challenge of multi-modality data input in LLMs by employing an AI-human cooperative approach for prompt generation, coupled with the use of a Recurrent Neural Network (RNN) for temporal representation learning of epidemiological time series. This strategy enables an LLM to effectively process complex pandemic-related information, encompassing spatial, epidemiological time series, public health policy, and genomic surveillance aspects. Furthermore, by recasting forecasting as an ordinal classification of hospitalization trends, we align the model's output with the needs of public health decision-makers, also adhering to CDC guidance \cite{cdc_tracker}. An extensive evaluation of PandemicLLM from all 50 U.S. states across 16 weeks demonstrates its advantages: PandemicLLM delivers \textbf{robust and trustworthy forecasts}, offering categorical predictions with confidence levels that support public health policy making. Moreover, it exhibits \textbf{timely and accurate response to emerging variants}, leveraging the textual analysis of timely genomic surveillance data. These findings highlight the potential of leveraging LLMs in public health emergency response settings.

\section{Novelties and contributions}

By reformulating pandemic forecasting as a text reasoning problem, PandemicLLMs push the boundaries of traditional forecasting methods. To the best of our knowledge, this study represents the first LLM-based modeling framework that incorporates novel data streams for forecasting pandemics and contributes to the literature by:


\begin{enumerate}
    \item \textbf{Extending the LLM framework for enhanced pandemic forecasting:} We extended the existing LLM architecture to handle the multi-modality nature of pandemic-related data. This extension includes integrating a human-AI collaborative prompt design, enabling the transformation of numerical pandemic data into textual formats suitable for LLM-based learning. Crucially, we introduced a RNN encoder tailored for the temporal representation learning of epidemiological time series, which, as our ablation study highlights, contributes to an accuracy improvement of 17\%-24\%.
    \item \textbf{Incorporating underutilized pandemic-related data streams into pandemic forecasting:} PandemicLLM incorporates critical and timely disease-relevant information and novel data streams that have not previously been used by pandemic forecasting models. Specifically, PandemicLLM integrates real-time textual virological characteristics, estimated variant prevalence, textual public health policy, and healthcare system performance alongside the traditional demographic and epidemiological time series. The inclusion of the variant information is shown to improve model performance without the need for retraining, underscoring the importance of adaptability to real-time information for effective response during ongoing outbreaks.

    \item \textbf{Providing public health decision-makers with robust and trustworthy predictions:} Our model is engineered to create clear and easily understood outputs. It effectively differentiates between highly certain outcomes and those with more uncertainty. Our findings indicate that as the confidence level of the predictions rises, so does the model's accuracy. This ensures that decision-makers have access to dependable forecasts, thus increasing the utility of the model.
\end{enumerate}

\section{Data and Methods}

\label{sec:results}
\subsection{Multi-modality pandemic data }
\label{sec:fig1a}

In addressing the intricate dynamics of COVID-19 transmission, which involves data of heterogeneous formats, the proposed \model was built on four data categories: \textbf{spatial}, \textbf{epidemiological time series}, \textbf{public health policy}, and \textbf{genomic surveillance data} (Fig. \ref{fig: fig_1}a). This categorization reflects the distinct nature and representation of each data type. Spatial data, sourced at the state level, comprised numeric and static variables, including demographic information, healthcare system scores, and political affiliations. Epidemiological time series were collected at weekly resolution for each state, encompassing numerical and sequential data, such as reported COVID-19 cases, hospitalization, and vaccination rates. Public health policy data, also state-specific, were captured in a textual format, detailing the stringency level and types of government policies. Genomic surveillance data are a hybrid of textual and sequential formats, with textual data sourcing from the authoritative reports of variants' virological characteristics and sequential data reflecting the current prevalence of these variants. 


Our multi-modality pandemic dataset encompasses information spanning 50 states across the U.S. and 104 weeks from January 2021 to January 2023, culminating in a total of 5,200 data records. This dataset includes five types of public health policies, three variations in vaccine rates, and genomic data on five Omicron sublineages. Utilizing the AI-human cooperative prompt design (section \ref{prompt_design}), these data records have been transformed into a set of prompts, aggregating approximately 1.51 million words. Further details regarding the data sources and the methods used in data preprocessing are documented in Section \ref{sec:datasets}. 

\subsection{The \ours framework}


To provide actionable and interpretable forecasts, we formulate pandemic forecasting as an ordinal classification problem. Leveraging the heterogeneous data types discussed in section \ref{sec:fig1a}, we adapt these data into a textual format amenable to LLM learning facilitated by an AI-human cooperative prompt design. Subsequently, the \model undergoes supervised fine-tuning using designed prompts and targets. To ensure the accuracy and reliability of our predictions, we undertake an extensive performance evaluation, focusing particularly on the results' trustworthiness. Fig. \ref{fig: fig_1}b illustrates the entire construction pipeline of \model, with subsequent sections providing detailed explanations of each segment of the pipeline. 


\subsubsection{Pandemic forecasting as ordinal classification}
\label{sec: ordinal_targets_design}

In our effort to build \model, particular attention was given to the design of the prediction targets, aiming to refine the targets used for COVID-19 forecasting.  We argue that continuous targets, such as reported cases or hospitalizations, are prone to reporting errors and hinder clear uncertainty communication for stakeholders. The frequent ambiguity observed in the conflicting predicted trends of 50\% and 95\% confidence intervals from the COVID-19 Forecast Hub \cite{nixon2022real} highlights this problem. To address this, we introduce a straightforward yet informative targets: Hospitalization Trend Category (\target). This target categorizes future hospitalization trends into five levels: Substantial Decrease, Moderate Decrease, Stable, Moderate Increase, and Substantial Increase. Detailed definitions for \target can be found in Section \ref{sec:temporal data}.

\subsubsection{AI-human cooperative prompt design}

\begin{figure}[h!]
    \includegraphics[width=1\textwidth]{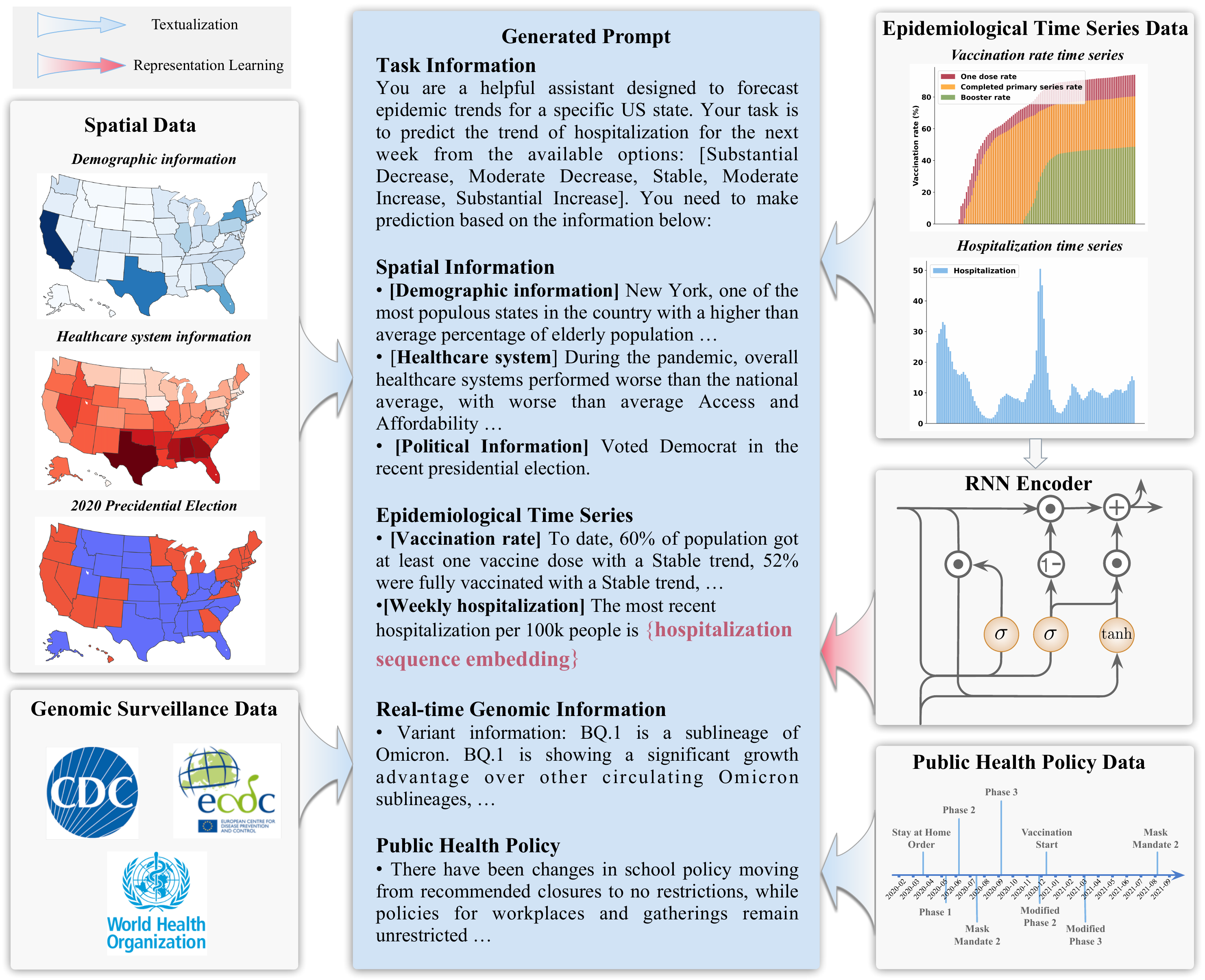}
    \caption{\textbf{Summary of the AI-human cooperative prompt design.} Spatial data for all 50 U.S. states are converted into descriptions to reflect their rankings; the policy data includes stringency levels and changes from week-to-week. Epidemiological time series data uses both narrative generation and representation learning. Genomic surveillance data combines textual summaries of variant characteristics with recent prevalence. The blue arrow indicates the information textualization, while the red arrow indicates the sequence representation learning. Each designed prompt has 296 to 322 words.}
    \label{fig: fig_2}
\end{figure}

The cornerstone of the \model framework is the AI-human cooperative prompt design, a process that blends human insight with AI efficiency. This framework involves the embedding of multi-modality data into textual formats, and each resulting prompt has around 300 words (see Extended Data Fig. \ref{fig: example_prompt} for an example of completed prompt). Summaries of the data transformation into text are provided below (see Fig. \ref{fig: fig_2}), with detailed descriptions available in Section \ref{prompt_design}.



\begin{itemize}
    \item \textbf{Spatial data:} Across all 50 U.S. states, each type of spatial data (e.g., population and healthcare system scores) was assigned a numerical rank and then categorized into one of five descriptive levels reflecting its relative position.
    \item \textbf{Public health policy data:} For every state and week, each policy is detailed by including its policy type, summarizing its stringency level, and highlighting any variations compared to the policy of the preceding week. The policies included in this study are detailed in Extended Data Table \ref{tab:policies}. 
    \item \textbf{Epidemiological time series data:} In the embedding of epidemiological time series, two distinct approaches are implemented: (1) The first method uses ChatGPT\cite{ChatGPT} to convert numerical sequence data into detailed textual narratives. This is achieved by instructing ChatGPT to analyze and summarize the recent trends and rate of changes within the given time series data as a one-sentence description (see Extended Data Fig.\ref{fig: AI_textualization}). (2) The second approach focuses on the most critical temporal data elements, specifically the hospitalization rates time series, which are tokenized through representation learning utilizing a Gated Recurrent Unit (GRU) framework (see Extended Data Fig.\ref{fig: model_framework}). 
    \item \textbf{Genomic surveillance data:} As outlined in section \ref{sec:fig1a}, genomic surveillance data were collected in a combination of textual and sequential formats. For the textual component, a summary is created from authoritative reports (refer to Supplementary Information section 1 for detailed data sources.), focusing on three key virological characteristics of the variant: infectiousness, severity, and resistance to immunity. For the sequential aspect of genomic surveillance data, a methodology similar to the first approach for temporal data is employed, where ChatGPT is used to summarize the trends and rate of changes in the recent variant proportion time series. 
\end{itemize}

\subsubsection{Experiment setup}


We fine-tuned LLaMA2\cite{Llama2}—a publicly accessible LLM by Meta—to predict the Hospitalization Trend Category (HTC) for the upcoming 1 and 3-week for each state. As illustrated in Fig. \ref{fig: fig_1}b, the model was fine-tuned to maximize the probability of predicted tokens corresponding to the target category. Our modeling framework included testing LLMs of three varying scales, with parameter sizes of 7 billion (7B), 13 billion (13B), and 70 billion (70B), to assess the potential enhancement in performance corresponding to increased model size. However, given the computational requirement of the 70B model might not be practical for all public health institutions, we discuss the results of \ours-70B only in section \ref{sec: model_comparison} and analyze the more accessible 7B and 13B models in other sections. For performance evaluation, we used the data from September 2022 through January 2023 as our test set. The prior period, spanning January 2021 to September 2022, was partitioned into training and validation sets with an 80/20 ratio.

\subsubsection{Evaluation of \model and reference models}
\textbf{Evaluation metrics:} As our proposed framework reformulates pandemic forecasting as an ordinal classification problem, we adopt the widely used accuracy and mean squared error (MSE) for evaluation. Nevertheless, these two metrics only evaluate the prediction with the largest probability, overlooking the prediction distribution. For a fine-grained evaluation, we propose to evaluate using Weighted MSE (WMSE) along with Brier Score\cite{rufibach2010use}, and Ranked Probability Score (RPS)\cite{leung2021real}. Specifically, the Brier Score quantifies the precision of probability estimations. The RPS penalizes predictions where the predicted probabilities deviate from the target category. The WMSE escalates its penalization in proportion to the divergence of the predicted probabilities from the target category. The detailed definition for each error metric is documented in section \ref{Evaluation}.

\textbf{Reference models:} We evaluated \models with five models currently used in pandemic forecasting, including four machine learning models: Gated Recurrent Unit (GRU)\cite{shahid2020predictions}, Long Short-Term Memory (LSTM)\cite{du2023incorporating}, Bidirectional Long Short-Term Memory (bi-LSTM)\cite{aung2023novel}, and AutoRegressive Integrated Moving Average (ARIMA)\cite{sahai2020arima}. These machine learning models were trained on numerical data, explicitly excluding contextual policy and genomic surveillance information. In addition to these machine learning models, we included a simple yet hard to beat heuristic-based baseline, PrevTrend\cite{cramer2022united}, which assumes that the predicted probability for each state is based on the distribution of states across various categories for the most recent observation. Detailed descriptions of reference models are presented in the Supplementary Information sections 4 and 5.

\section{Results}
\label{overall_performance}

\subsection{COVID-19 hospitalization trend prediction}
\textbf{PandemicLLMs capture the overall COVID-19 hospitalizations trend accurately.} Designed to predict 1-week and 3-week COVID-19 Hospitalization Trend Category (HTC) for each U.S. state, PandemicLLMs' predictions closely align with the observed ground truth pattern (Fig.\ref{fig: prediction_with_target}). This pattern reveals a shared decline in hospitalizations throughout September and early October, followed by a distinct increase starting in November. Despite this shared trend, individual states exhibit highly diverse hospitalization trajectories, underscoring the inherent complexity of pandemic forecasting. Remarkably, even within this diverse landscape, both versions of PandemicLLM (7B and 13B) demonstrate similar performances, with accuracies of 55.4\% and 56\% for 1-week predictions and 45.4\% and 46.4\% for 3-week predictions, respectively.


\subsection{Spatial performance evaluation}
\label{sec: spatial_performance}
\textbf{\models exhibit robust performance nationally, though with local variations.} The heterogeneity in state-level hospitalization trajectories motivates the investigation of how PandemicLLMs' performance varies across states. Fig.\ref{fig:map_a} to \ref{fig:map_d} display the average performance on WMSE for each state, highlighting the spatial differences in model efficacy. Nationally, the PandemicLLM-7B and PandemicLLM-13B show average WMSE of 0.72 and 0.9 for 1 and 3-week forecasts, respectively. However, local variation still exists, given the state-level pattern diversity. Iowa (0.31, 0.49), Utah (0.48, 0.41), and Idaho (0.41, 0.53) exhibit the lowest WMSE (1-week, 3-week), indicating the top performances. Conversely, Delaware (1.46, 1.69), Wyoming (1.32, 1.53), and Vermont (1.31, 1.49) have the highest WMSE (1-week, 3-week), indicating the bottom performances. Based on our evaluation, the \model demonstrates reliable performance in the West Coast, Southeast, and the Great Lakes regions. This pattern likely stems from the similarity in hospitalization trends among states within these regions (see Fig. \ref{fig: prediction_with_target} for categorical trend and Extended Data Fig.\ref{fig: hosp_trend} for continuous trend). Notably, these regional trends closely align with the overall national pattern. For Northeastern and Midwestern regions, the regional behavioral differences\cite{zang2021us} potentially triggered the trend variation, leading to performance decreases in states with trends diverging significantly from the national trajectory (e.g., Wyoming, Maine, and South Dakota). This finding suggests that region-specific models could be a valuable avenue for future research to address these localized variations more effectively. 

\newpage
\begin{figure}[h!]
  \begin{subfigure}{1\textwidth}
  \centerline{\includegraphics[width=0.95\linewidth,height=0.5\linewidth]{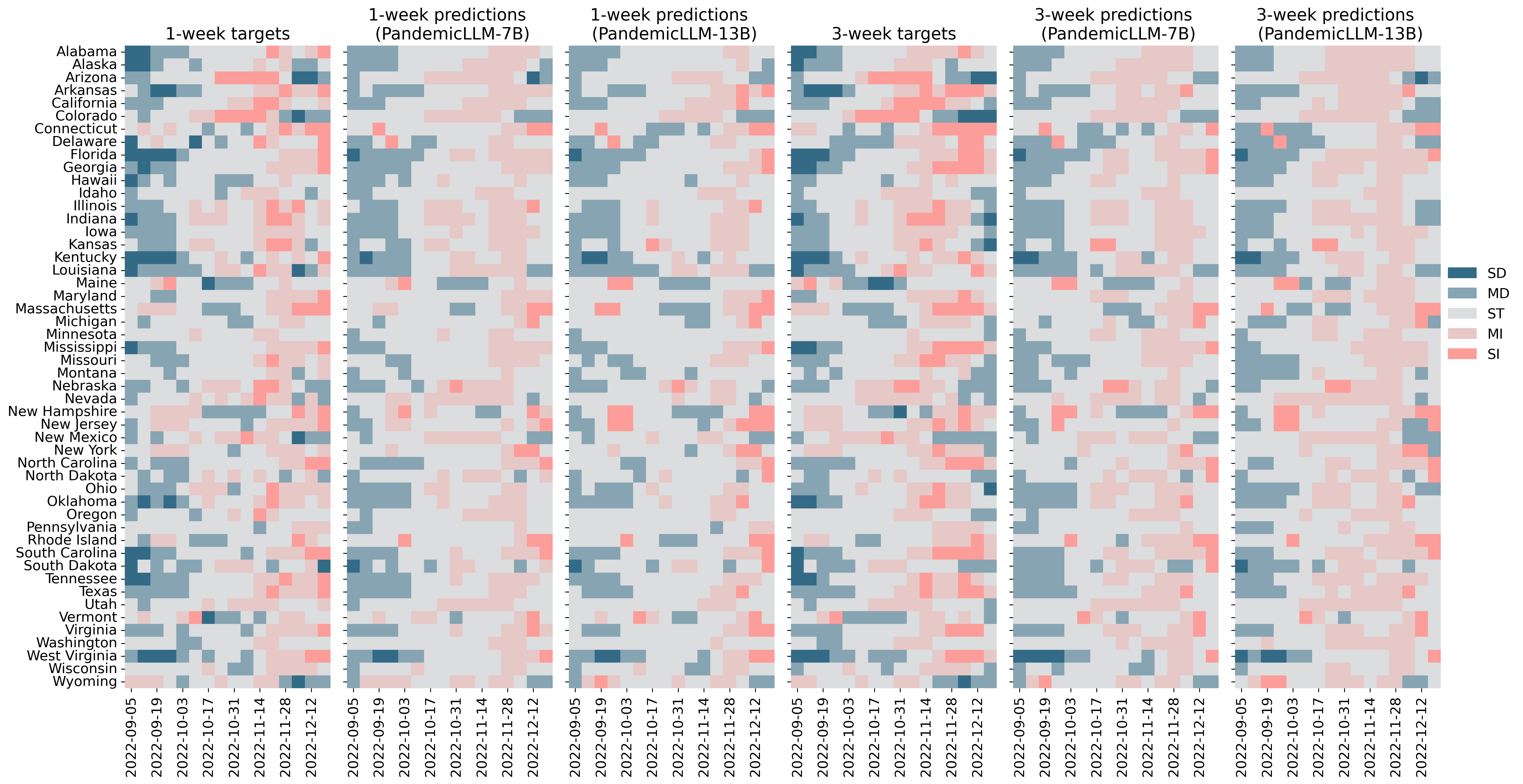}}
  \caption{Predictions visualization: 1-week and 3-week predictions vsiualization by PandemicLLMs versus the ground truth targets, in 50 States from September 5, 2022 to December 12, 2022. Color indicates Hospitalization Trend Category (HTC): SD: Substantial Decrease, MD: Moderate Decrease, ST: Stable, MI: Moderate Increase, SI: Substantial Increase.}
  \label{fig: prediction_with_target}
  \end{subfigure}
  
  \bigskip

  \begin{subfigure}{.5\textwidth}
  \centering
  \includegraphics[width=.95\linewidth]{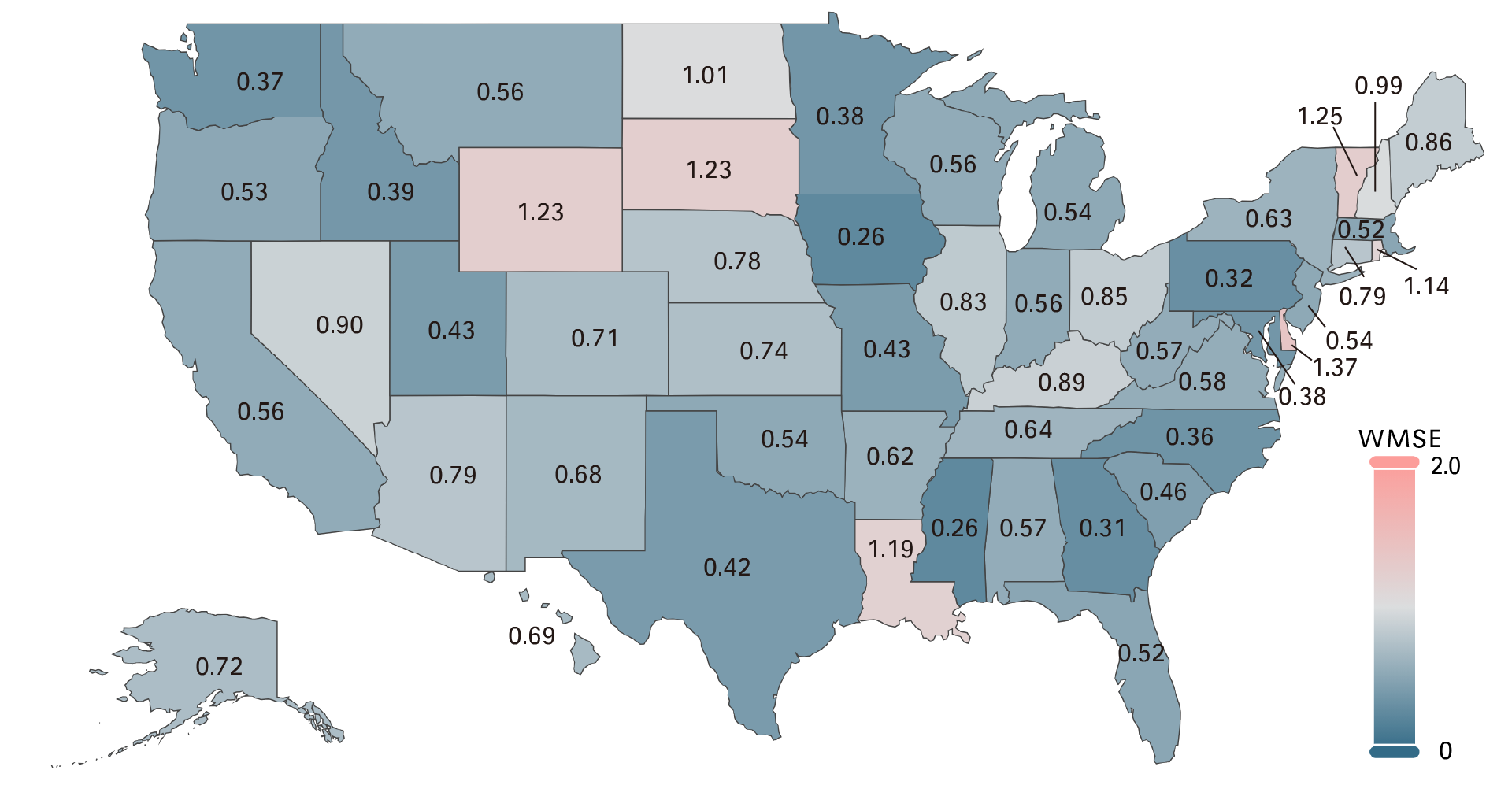}
  \caption{\model-7B performance by state (1-week)}
  \label{fig:map_a}
  \end{subfigure}%
  \begin{subfigure}{.5\textwidth}
  \centering
  \includegraphics[width=.95\linewidth]{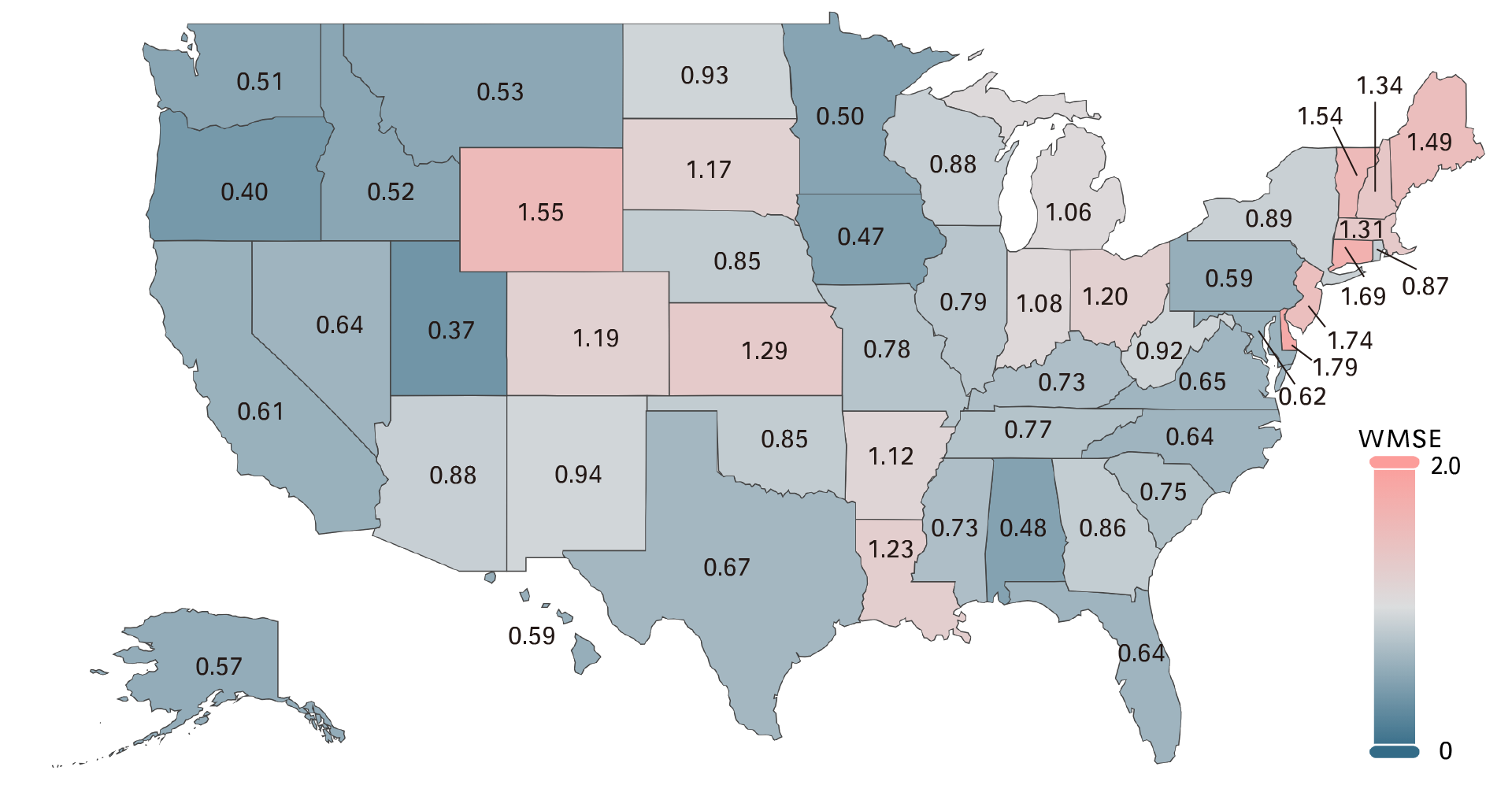}
  \caption{\model-7B performance by state (3-week)}
  \label{fig:map_b}
  \end{subfigure}

  \bigskip
  \begin{subfigure}{.5\textwidth}
  \centering
  \includegraphics[width=.95\linewidth]{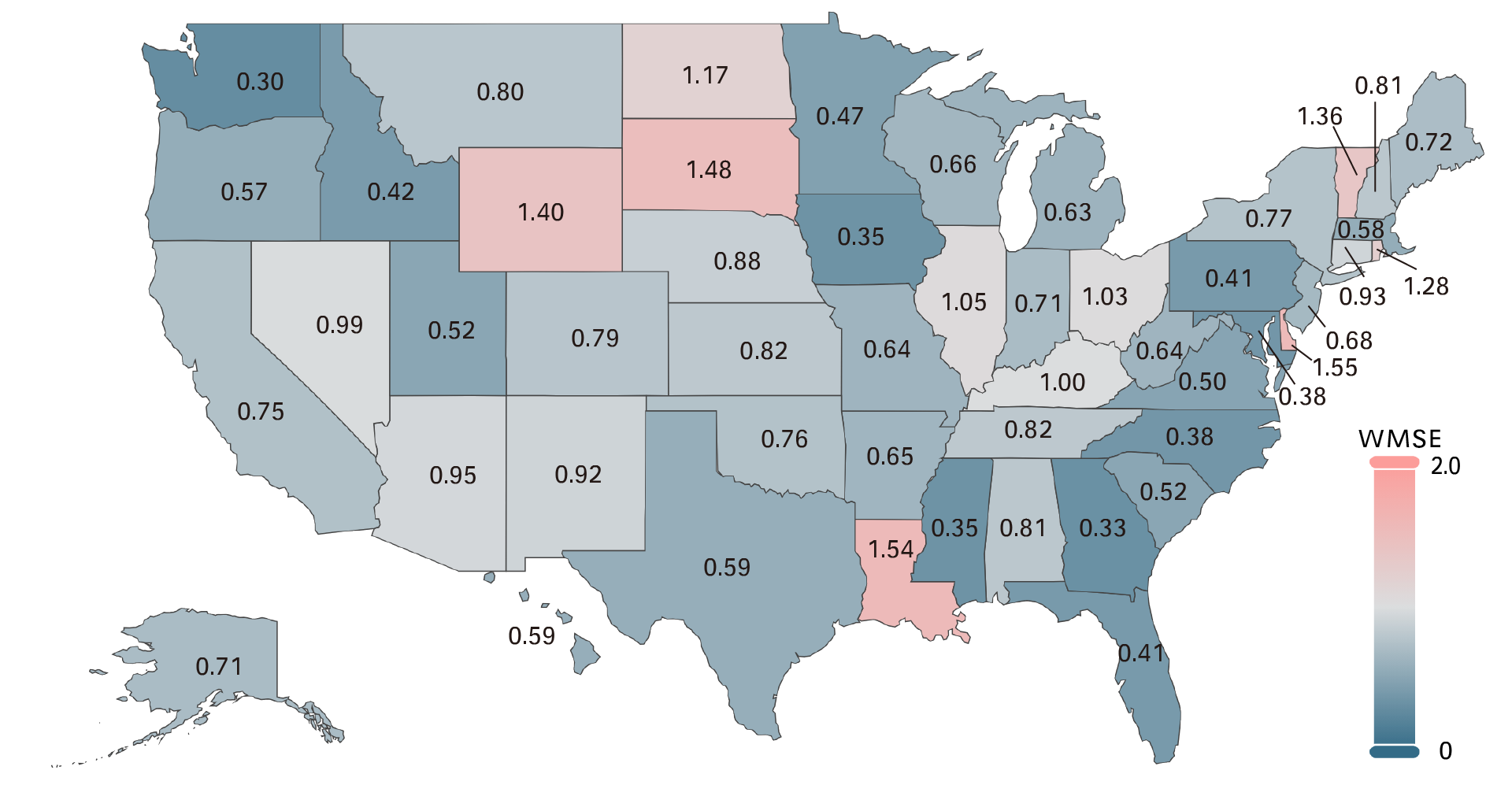}
  \caption{\model-13B performance by state (1-week)}
  \label{fig:map_c}
  \end{subfigure}%
  \begin{subfigure}{.5\textwidth}
  \centering
  \includegraphics[width=.95\linewidth]{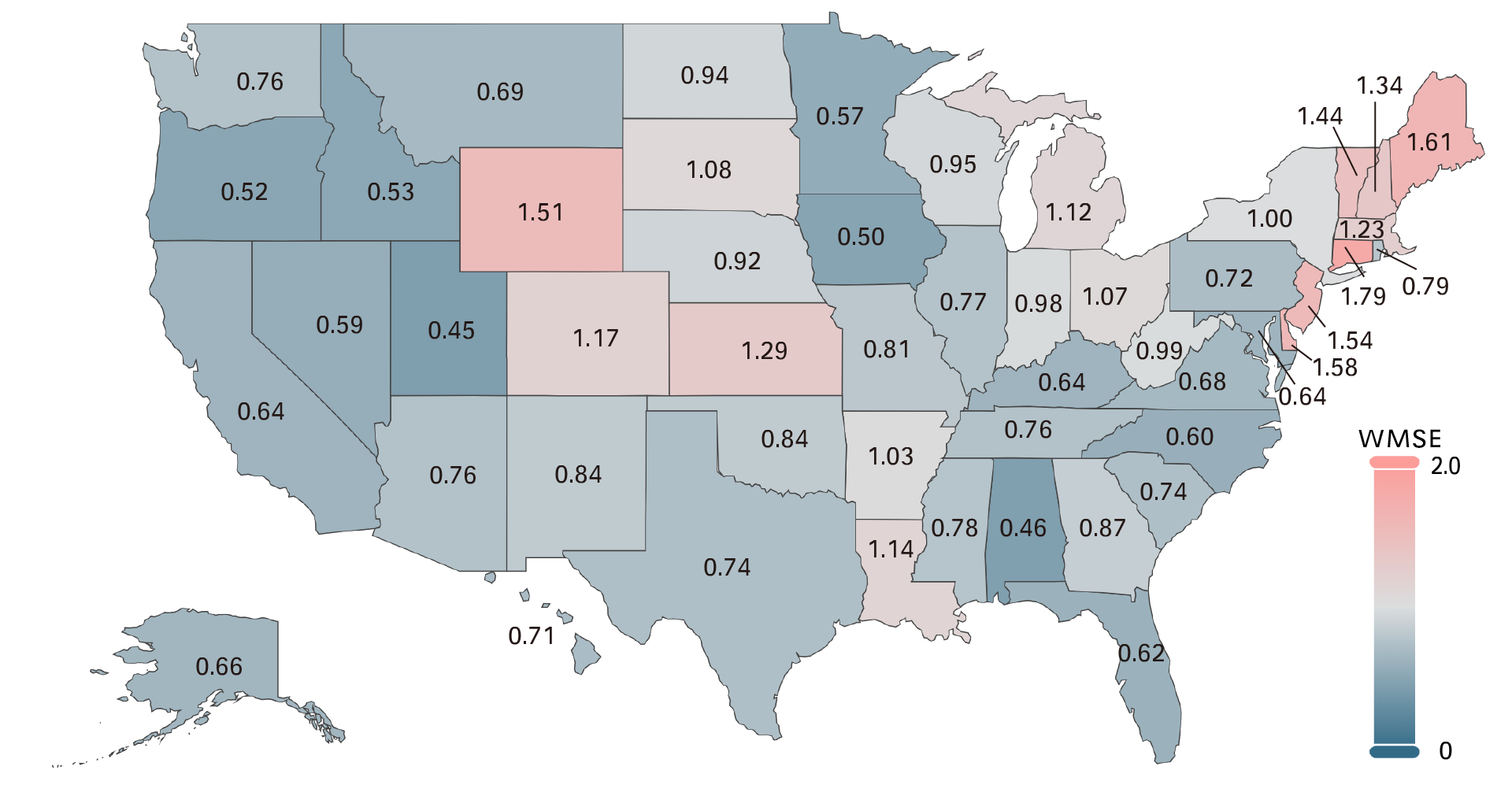}
  \caption{\model-13B performance by state (3-week)}
  \label{fig:map_d}
\end{subfigure}
\end{figure}

\begin{figure}\ContinuedFloat
\centering
\begin{subfigure}{1\textwidth}
\includegraphics[width=.95\linewidth]{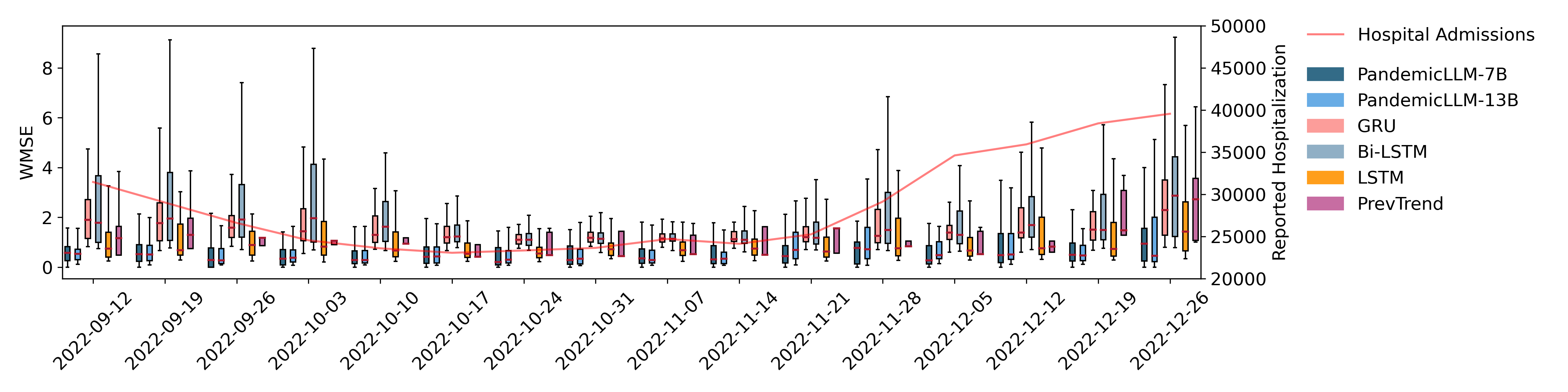}
\caption{1-week forecasting performance measured by weighted mean squared error (WMSE) over time.}
\label{fig:temporal_evaluation_week1}
\end{subfigure}

\bigskip
\begin{subfigure}{1\textwidth}
\includegraphics[width=.95\linewidth]{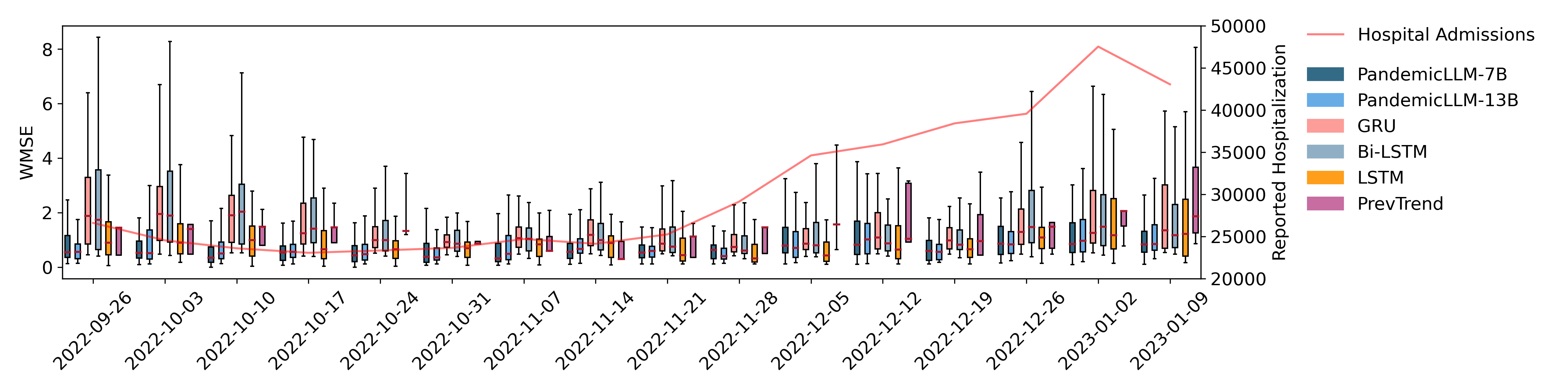}
\caption{3-week forecasting performance measured by weighted mean squared error (WMSE) over time.}
\label{fig:temporal_evaluation_week3}
\end{subfigure}
\caption{\textbf{\models' predictions visualization and performance evaluation.} \textbf{(a)}, 1-week and 3-week predictions by PandemicLLMs versus the ground truth targets. Color indicates Hospitalization Trend Category (HTC): SD: Substantial Decrease, MD: Moderate Decrease, ST: Stable, MI: Moderate Increase, SI: Substantial Increase. \textbf{(b, c)}, 1-week and 3-week performance for \model-7B. \textbf{(d, e)}, 1-week and 3-week performance for \model-13B. The color gradients represent the magnitude of the WMSE, where a darker shade of red signifies a greater error, and a darker shade of blue denotes a smaller error. Equivalent evaluations with alternative error metrics are included in the Supplementary Information section 9. \textbf{(f, g)}, Performances comparison of \models with reference models across time. The red curve on the back represents the weekly reported COVID-19 hospital admission at the national level. The left y-axis represents the scale of WMSE, and the right y-axis represents the scale of hospital admission. Each set of bar graphs in the figure represents the distribution of \metric for all states during a specific week. The color bars represent the error distributions for different models. \textbf{(f)}, 1-week forecasting performance. \textbf{(g)}, 3-week forecasting performance.}
\label{fig: model all}
\end{figure}

\subsection{Comparison to reference models}

In light of the observed spatial variations in PandemicLLMs' performance, and to comprehensively assess its capabilities, this section presents a comparative analysis of the PandemicLLMs' overall performances against four established machine learning models and one heuristic-based baseline. Utilizing five metrics, including Accuracy, MSE, WMSE, Brier Score, and RPS, for evaluation, the average performances across all states and weeks tested are shown in Table~\ref{tab:performance:covid}, from which we have the following observations:

\textbf{\models outperform existing forecasting models by at least 20\%.} The heuristic-based benchmark PrevTrend has average rankings of 4th and 6th for 1-week and 3-week prediction, demonstrating the complex nature of pandemic forecasting, which poses significant challenges for traditional machine-learning models. However, with the ability to leverage multi-modality data, all sizes of \models demonstrate better average rankings compared to all reference models, leading to average accuracy improvements of at least 20\% and 22\% for 1-week and 3-week, respectively. The notable performance enhancement validates the effectiveness of our proposed framework.

\label{sec: model_comparison}
\begin{table}[H]
    \small
    \begin{center}
    \caption{A summary of overall models' performances for \models, baseline, and other machine learning models. $\uparrow$/$\downarrow$ indicate that higher/lower metric values signify better performance. Numbers in parentheses represent the relative ranking of each model.}
    \label{tab:performance:covid}
    \setlength{\tabcolsep}{4pt} 
    \renewcommand{\arraystretch}{1.1} 
    \resizebox{0.9\textwidth}{!}{
    \begin{tabular}{c|c|c|c|c|c|c|c}
    \toprule[1.5pt]
     \multirow{2}{*}{\makecell[c]{\textbf{Prediction Target}}} & \multirow{2}{*}{\textbf{Model}} & \multicolumn{5}{c}{\textbf{Evaluation Metric (Model Rank)}} \vline & \multirow{2}{*}{\makecell[c]{\textbf{Average Rank}}}\\
     \cline{3-7}
      & & \textbf{Accuracy $\uparrow$} & \textbf{MSE $\downarrow$} & \textbf{WMSE $\downarrow$}  & \textbf{Brier Score $\downarrow$} & \textbf{RPS$\downarrow$} 
      \\
      
      \hline
      
\multirow{8}{*}{1-week}
& GRU \cite{} & 0.468 (5) & 0.899 (5) & 1.727 (6) & 0.667 (3) & 0.112 (4) & 4.6 (5)\\
& Bi-LSTM \cite{} & 0.365 (8) & 1.532 (8) & 2.134 (7) & 0.750 (7) & 0.154 (7) & 7.4 (8)\\
& LSTM \cite{} & 0.419 (6) & 0.896 (4) & 1.136 (4) & 0.747 (6) & 0.126 (6) & 5.2 (6)\\
& ARIMA \cite{} & 0.416 (7) & 1.108 (7) & \textbackslash{} & \textbackslash{} & \textbackslash{} & 7.0 (7)\\
& PrevTrend \cite{} & 0.471 (4) & 0.925 (6) & 1.361 (5) & 0.660 (2) & 0.113 (5) & 4.4 (4)\\
\cline{2-8} 
& \model-7B& 0.554 (3) & 0.593 (2) & 0.668 (2) & 0.668 (5) & 0.098 (2) & 2.8 (3)\\
& \model-13B & 0.560 (2) & 0.627 (3) & 0.767 (3) & \textbf{0.634} (1) & 0.098 (2) & 2.2 (2)\\
& \model-70B & \textbf{0.571} (1) & \textbf{0.560} (1) & \textbf{0.638} (1) & 0.667 (3) & \textbf{0.097} (1) & \textbf{1.4}(1)\\
 \cline{1-8}
  
\multirow{8}{*}{3-week}
& GRU \cite{} & 0.378 (4) & 1.067 (5) & 1.576 (5) & 0.745 (4) & 0.138 (4) &  4.4 (4)\\
& Bi-LSTM \cite{} & 0.369 (5) & 1.255 (7) & 1.678 (7) & 0.767 (6) & 0.151 (7) & 6.4 (7) \\
& LSTM \cite{} & 0.362 (7) & 0.936 (4) & 1.057 (4) & 0.923 (7) & 0.147 (6) &  5.6 (5)\\
& ARIMA \cite{} & 0.367 (6) & 1.308 (8) & \textbackslash{} & \textbackslash{} & \textbackslash{} & 7.0 (8)\\
& PrevTrend \cite{} & 0.343 (8) & 1.201 (6) & 1.588 (6) & 0.761 (5) & 0.140 (5) & 6.0 (6)\\
\cline{2-8} 
& \model-7B & 0.454 (3) & 0.797 (2) & \textbf{0.899} (1) & 0.739 (3) & 0.114 (3) & 2.4 (3)\\
& \model-13B& 0.464 (2) & \textbf{0.760} (1) & 0.908 (2) & 0.695 (2) & \textbf{0.106} (1) & \textbf{1.6} (1)\\
& \model-70B & \textbf{0.486} (1) & 0.805 (3) & 0.948 (3) & \textbf{0.687} (1) & 0.110 (2) & 2.0(2)\\
    \bottomrule[1.5pt]
    \end{tabular}
    }
    \end{center}
\end{table}

      
  
      


\textbf{Scaling the parameter size of the PandemicLLM could lead to performance improvements.} To evaluate the impact of parameter size on performance, the 70B version is included specifically in this section. For the 1-week forecast, PandemicLLMs with 7B and 13B parameters shown similar performance, whereas the 70B version leads in average rank and surpasses the others in all error metrics except Brier Score. Specifically, the 70B model improves Accuracy by 6.4\%, MSE by 17.1\%, WMSE by 13.8\%, and RPS by 3\%. For 3-week forecasts, the differences among the PandemicLLMs sizes are less distinct with 13B and 70B achieving both better results compared with 7B. While this analysis sheds some light on the possible impact of increased model complexity on model performance, this area needs further research.

The notable performance enhancement of PandemicLLM underscores its potential for robust pandemic forecasting.  A critical aspect of real-world forecasting is the ability to adapt to changing disease dynamics over time. Accordingly, the following analysis investigates how PandemicLLM's performance varies across different time periods.

\label{sec: temporal_performance}
\textbf{\models show robust performances across time.} Fig.\ref{fig:temporal_evaluation_week1} to \ref{fig:temporal_evaluation_week3} illustrate a comparative analysis of \models' performance relative to the reference models across the 16-week periods evaluated, where the national hospitalization trend changed from decreasing to stable and then to increasing. Each bar plot represents the WMSE distribution across all states for a specific week, where a lower bar indicates a better performance. This analysis reveals two primary findings: 1) The \models consistently outperform other models, exhibiting the most minor variability over time. 2) The \models show the most significant performance improvement, particularly as the outbreak exhibits a decreasing trend or approaches a peak. The \models' temporal performance highlights their adaptability to evolving disease dynamics, demonstrating effectiveness across various outbreak stages. Equivalent evaluations with alternative error metrics are included in the Supplementary Information sections 8.

\subsection{Trustworthy and robust results}
\label{sec:Trustworthiness}

\begin{figure}[h!]
\begin{subfigure}{.5\textwidth}
  \centering
  \includegraphics[width=.95\linewidth]{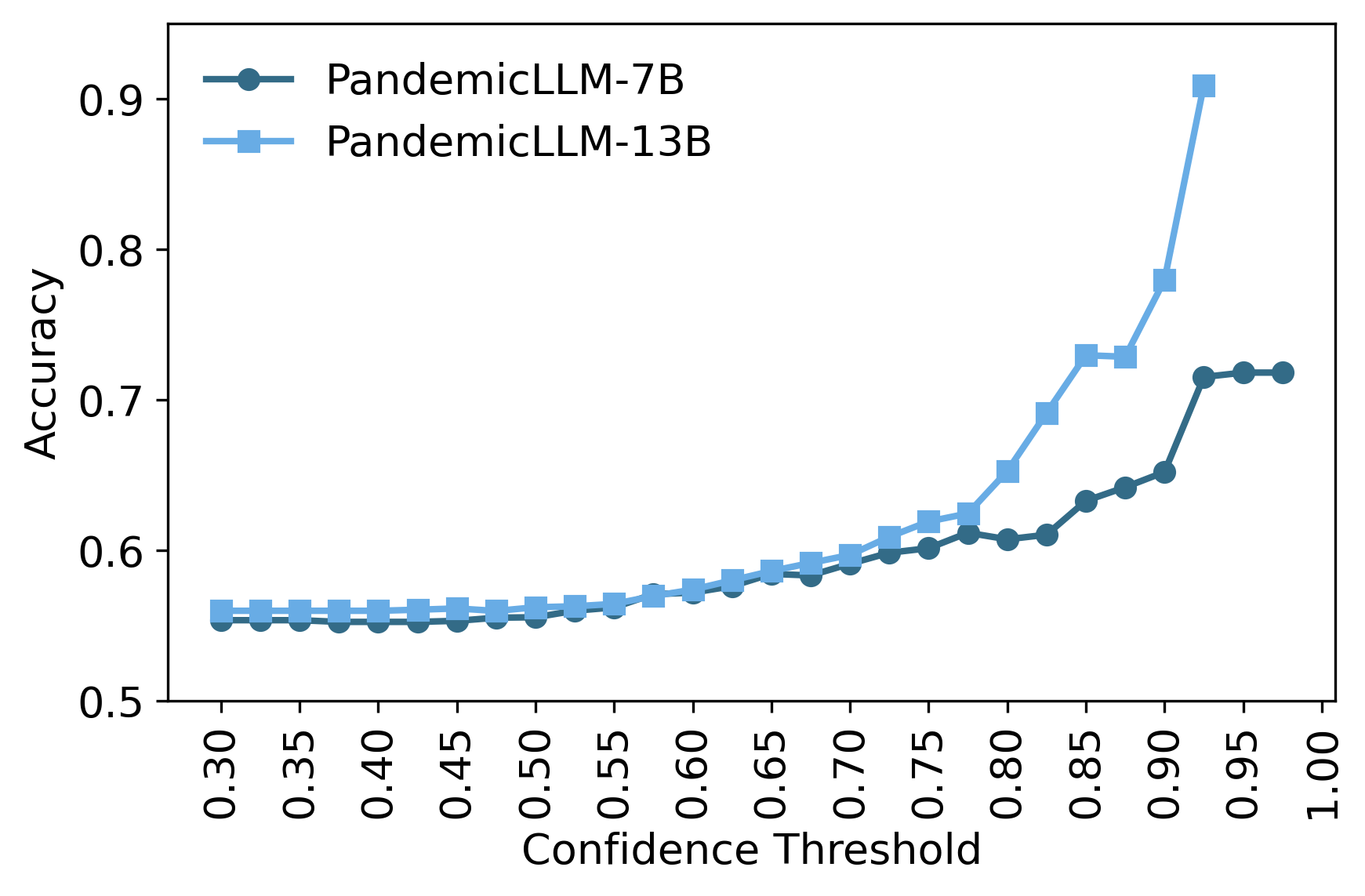}
  \caption{Confidence-Accuracy relation \textbf{(1-week)}}
  \label{fig:trustworthy_a}
\end{subfigure}%
\begin{subfigure}{.5\textwidth}
  \centering
  \includegraphics[width=.95\linewidth]{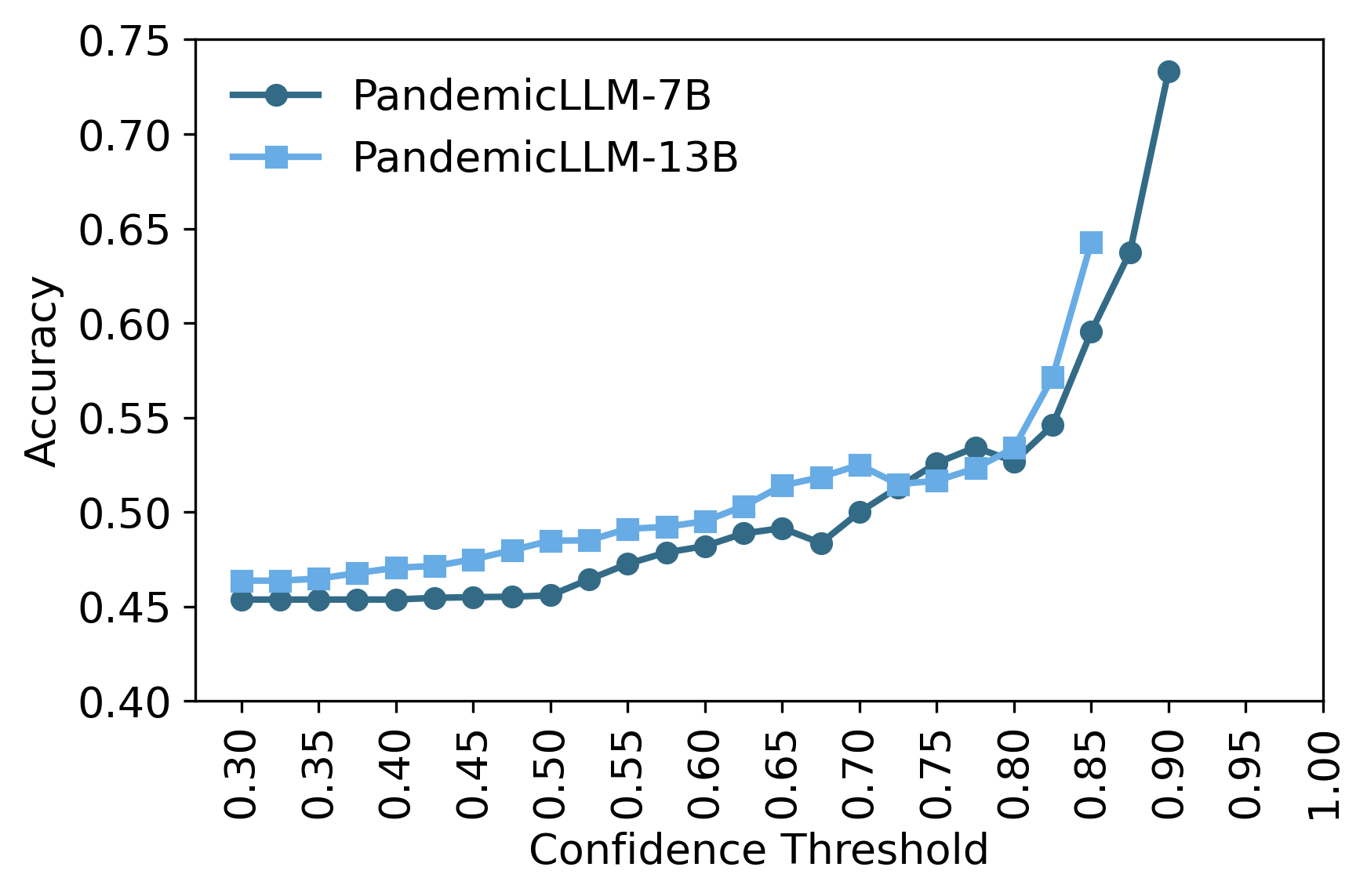}
  \caption{Confidence-Accuracy relation \textbf{(3-week)}}
  \label{fig:trustworthy_b}
\end{subfigure}

\bigskip
\begin{subfigure}{.5\textwidth}
  \centering
  \includegraphics[width=0.95\linewidth]{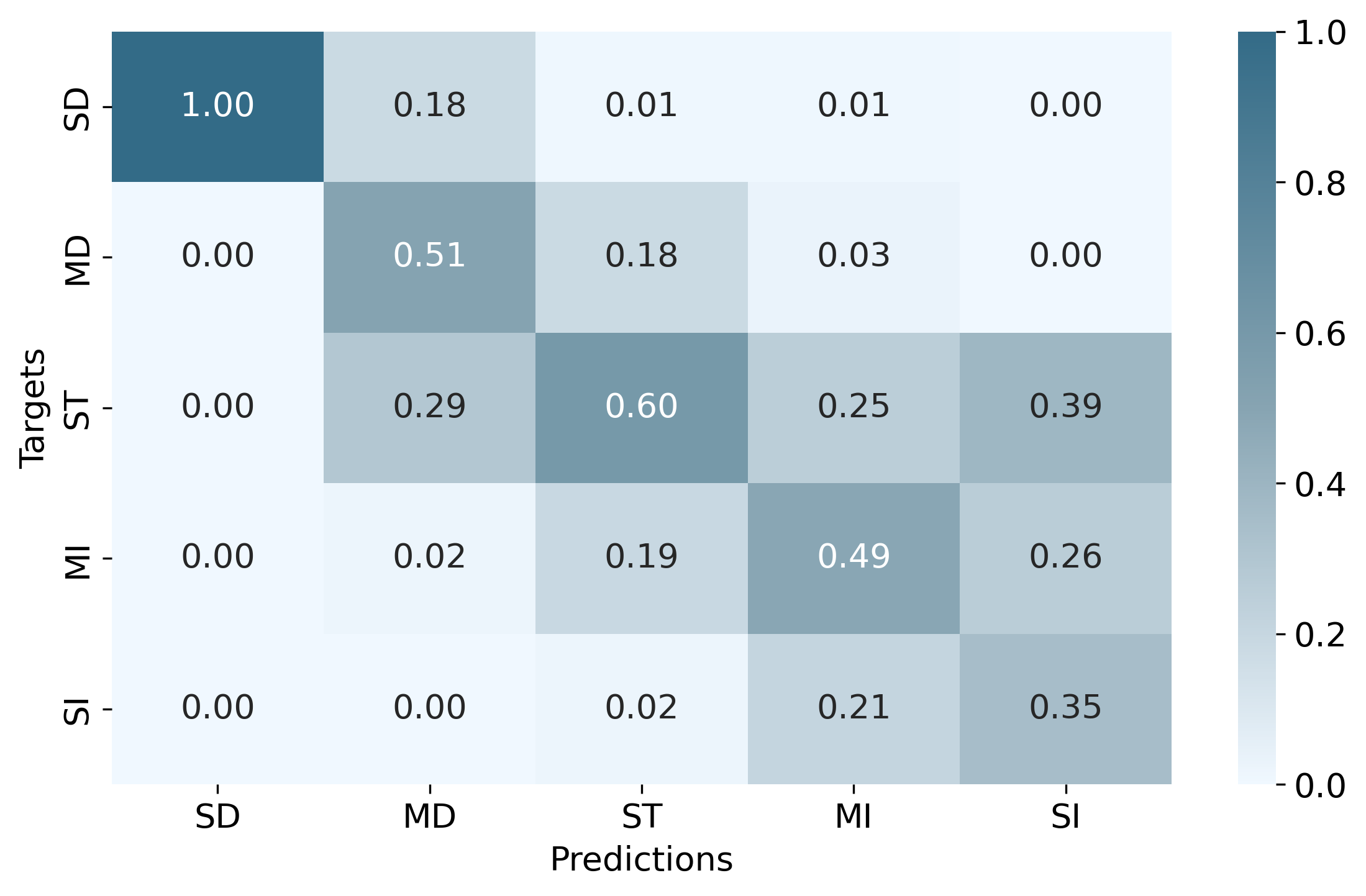}
  \caption{\model-\textbf{7B} confusion matrix \textbf{(1-week)}}
  \label{fig:trustworthy_c}
\end{subfigure}%
\begin{subfigure}{.5\textwidth}
  \centering
  \includegraphics[width=0.95\linewidth]{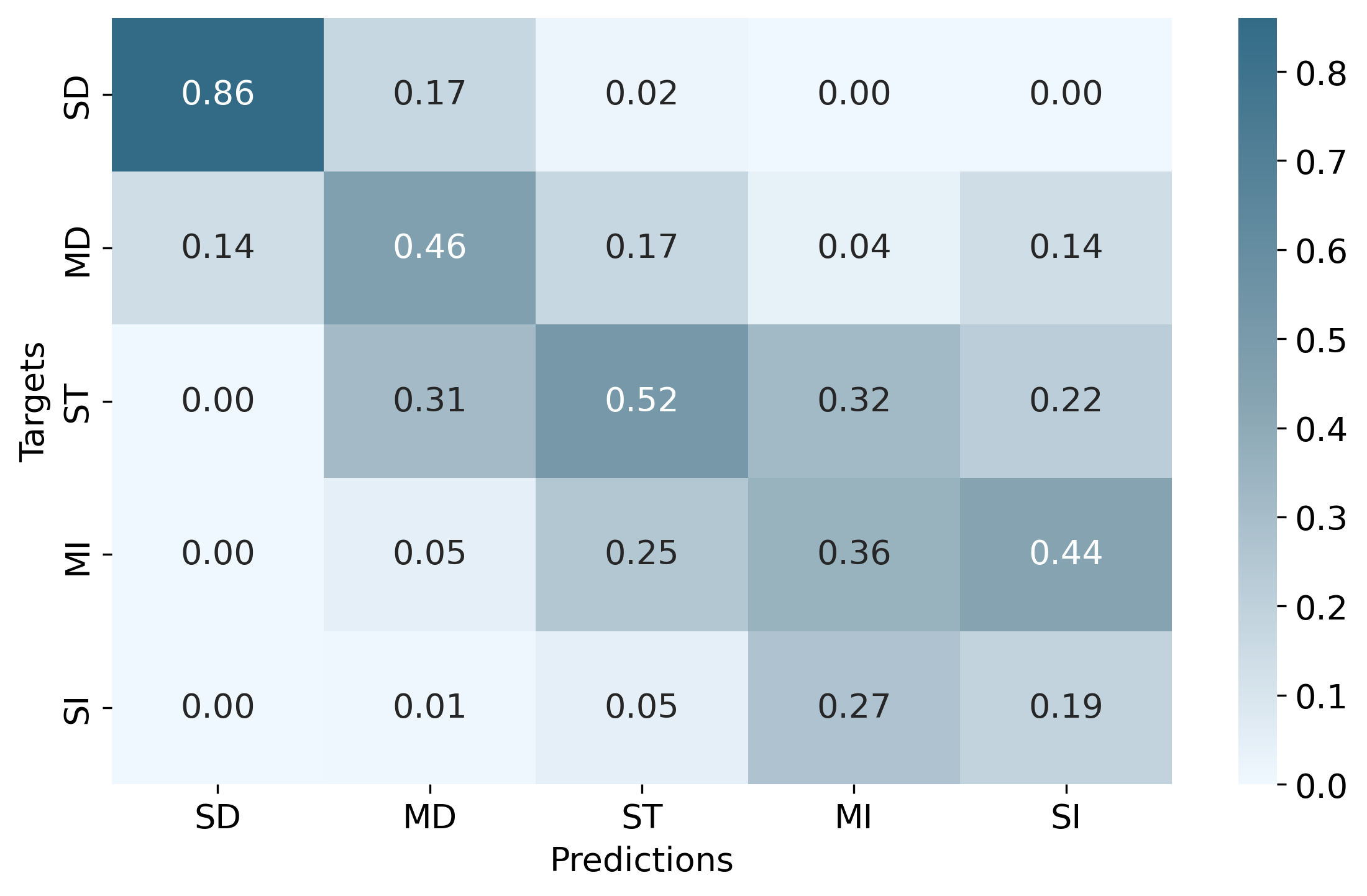}
  \caption{\model-\textbf{7B} confusion matrix \textbf{(3-week)}}
  \label{fig:trustworthy_d}
\end{subfigure}

\bigskip
\begin{subfigure}{.5\textwidth}
  \centering
  \includegraphics[width=.95\linewidth]{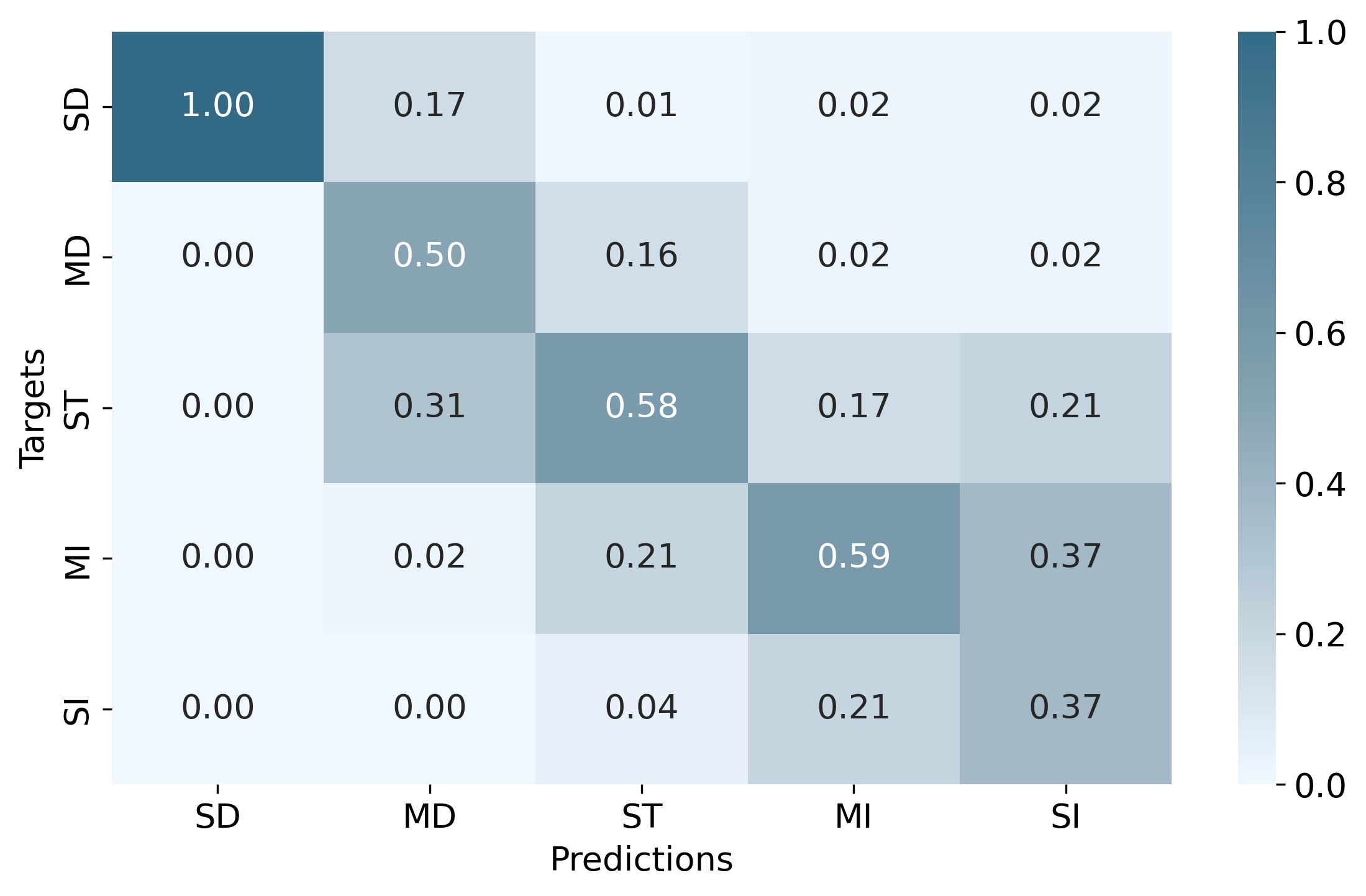}
  \caption{\model-\textbf{13B} confusion matrix \textbf{(1-week)}}
  \label{fig:trustworthy_e}
\end{subfigure}%
\begin{subfigure}{.5\textwidth}
  \centering
  \includegraphics[width=.95\linewidth]{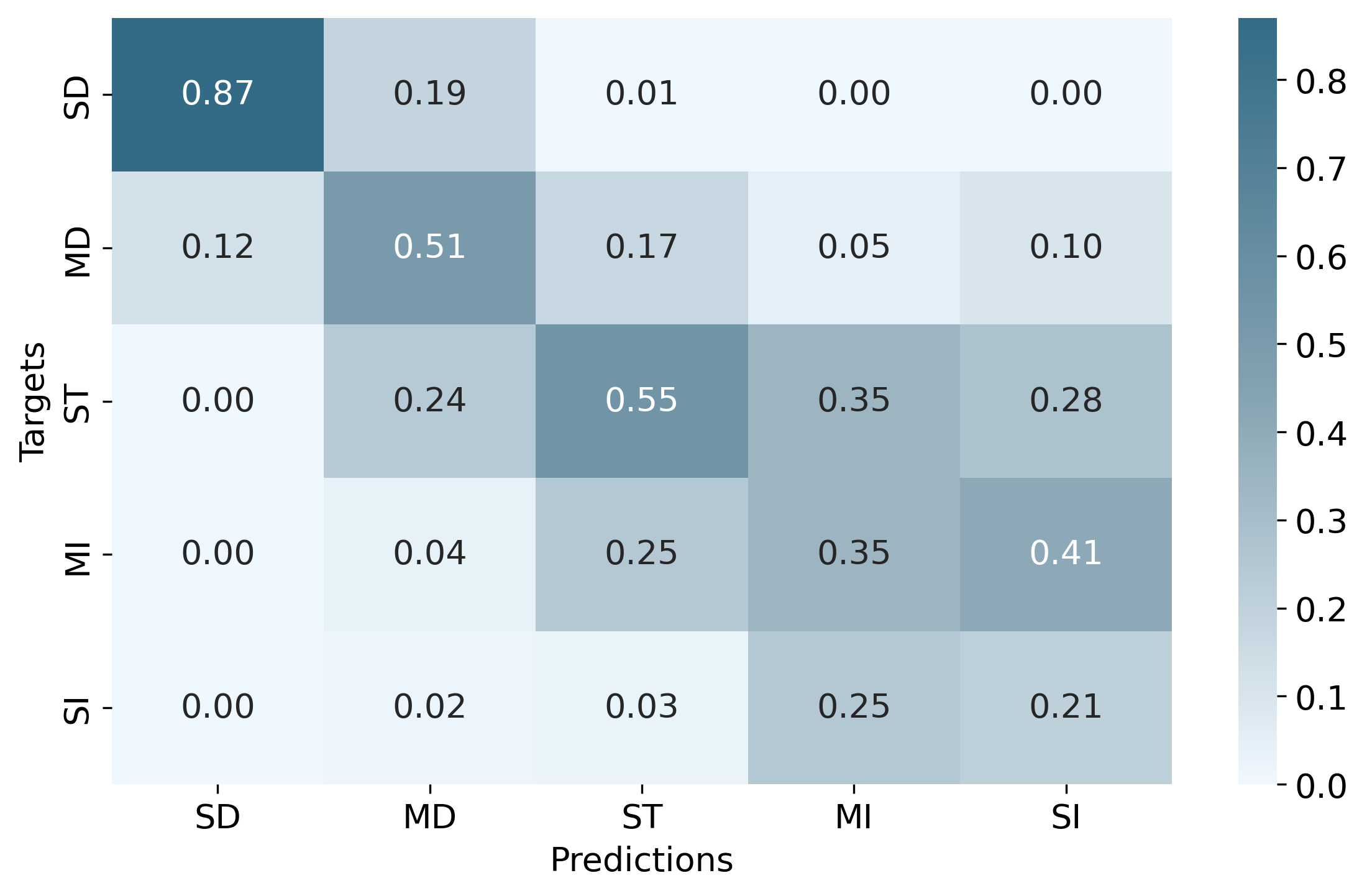}
  \caption{\model-\textbf{13B} confusion matrix \textbf{(3-week)}}
  \label{fig:trustworthy_f}
\end{subfigure}

\caption{\textbf{Trustworthiness for \models.} \textbf{ (a, b)} The accuracy of 1-week and 3-week predictions varied across various levels of prediction confidence. \textbf{(c, d)} The 1-week and 3-week confusion matrix for \model-7B. \textbf{(e, f)} The 1-week and 3-week precision confusion matrix for \model-13B. SD: Substantial Decrease, MD: Moderate Decrease, ST: Stable, MI: Moderate Increase, SI: Substantial Increase.}
\end{figure}

\textbf{High-confidence predictions demand real action.} The standard for models designed to aid public health decision-making is rigorous, necessitating a reliable model that provides clear guidance on its utility. In response, we show that the confidence level of \models, the highest probability assigned to the predicted category, is a robust indicator of reliability. Specifically, we present the models' accuracy with respect to different confidence thresholds in Figure \ref{fig:trustworthy_a} and \ref{fig:trustworthy_b}. As we elevate the confidence threshold, wherein only predictions surpassing this criterion are considered, notable enhancements in prediction accuracy are observed for both the 7B and 13B models. For instance, setting the confidence threshold at 0.85, our 13B model attains an accuracy of 73\% for 1-week forecasts and 64\% for 3-week forecasts. This characteristic of our model is invaluable for public health policy decision-making, offering decision-makers the flexibility to devise strategies grounded in the predictive confidence provided by the model.

\textbf{Dependable forecasts for informed public health decisions.} The stakes of making incorrect public health decisions are exceedingly high, such as prematurely easing restrictions when future hospital admissions are rising. A reliable forecasting model is paramount to prevent significant prediction errors, for instance, misidentifying a "Substantial Increase" as a "Substantial Decrease," which could severely undermine public health strategies. To evaluate the PandemicLLM's reliability under different pandemic phases, we analyze the confusion matrices displayed in Fig.~\ref{fig:trustworthy_c} to~\ref{fig:trustworthy_f}.  The findings underscore two pivotal insights: (1) The PandemicLLM exhibits strong predictive capability during the decreasing phases, particularly in identifying substantial decreases. This attribute is invaluable for informing policies on reopening and easing restrictions safely. (2) Errors in forecasting predominantly occur between neighboring categories. For instance, when our 13B model forecasts ``Substantial Increase'' predictions, the actual situation reflected at least a ``Moderate Increase'' in 74\% of 1-week forecasts and in 62\% for 3-week forecasts. This level of reliability suggests that PandemicLLM can offer valuable insights for making informed operational decisions across various phases of a pandemic.

\subsection{Integrating real-time genomic surveillance information for timely response}
\label{sec: variant}
One unique ability of \model is that it can incorporate previously unseen information through text reasoning, which allows the integration of timely pandemic-related information. One notable example is the emergence of new variants that pose significant challenges for traditional real-time forecasting. In this section, we present the capability of our model to mitigate such challenges by incorporating timely information on new variants. This analysis focuses on 3-week forecasting, aiming to provide an extended lead time for informed decision-making. 

\begin{figure}[h!]
\begin{subfigure}{1\textwidth}
  \centering
  \includegraphics[width=.9\linewidth]{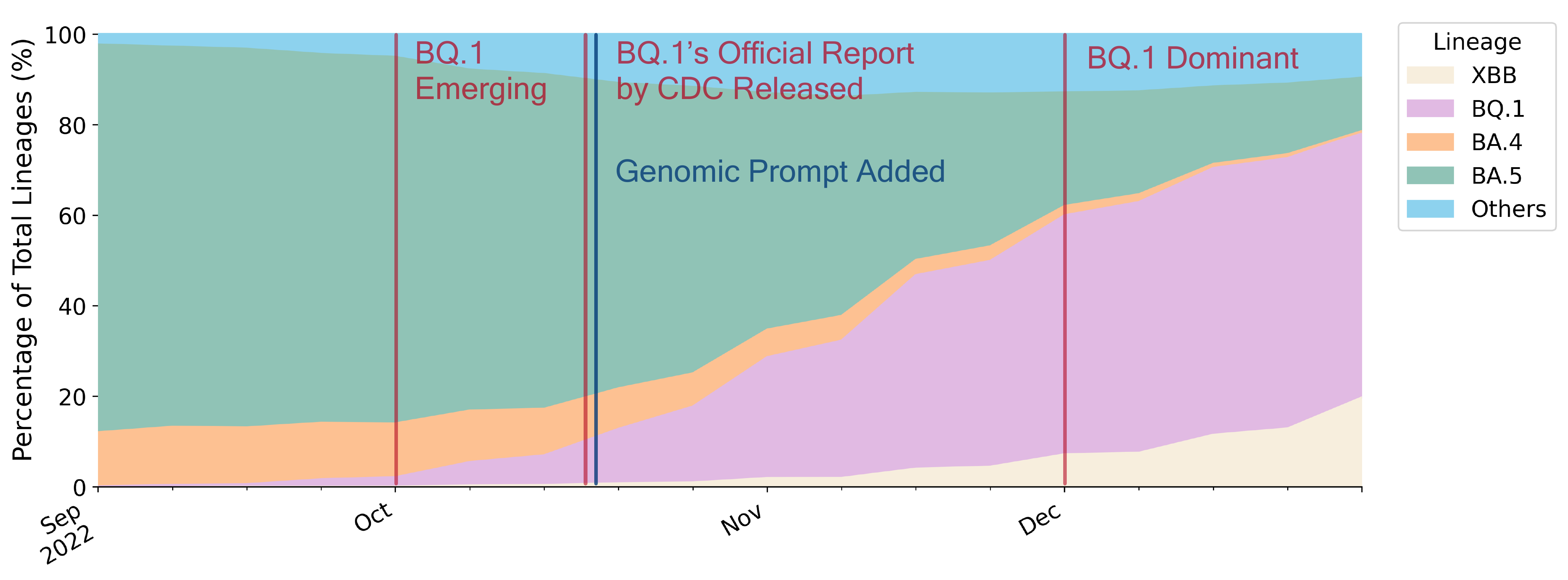}
  \caption{Proportional distribution of SARS-CoV-2 variants from September 2022 to January 2023.}
  \label{fig:variant_a}
\end{subfigure}%

\bigskip

\begin{subfigure}{.5\textwidth}
  \centering
  \includegraphics[width=.95\linewidth,height=.6\textwidth]{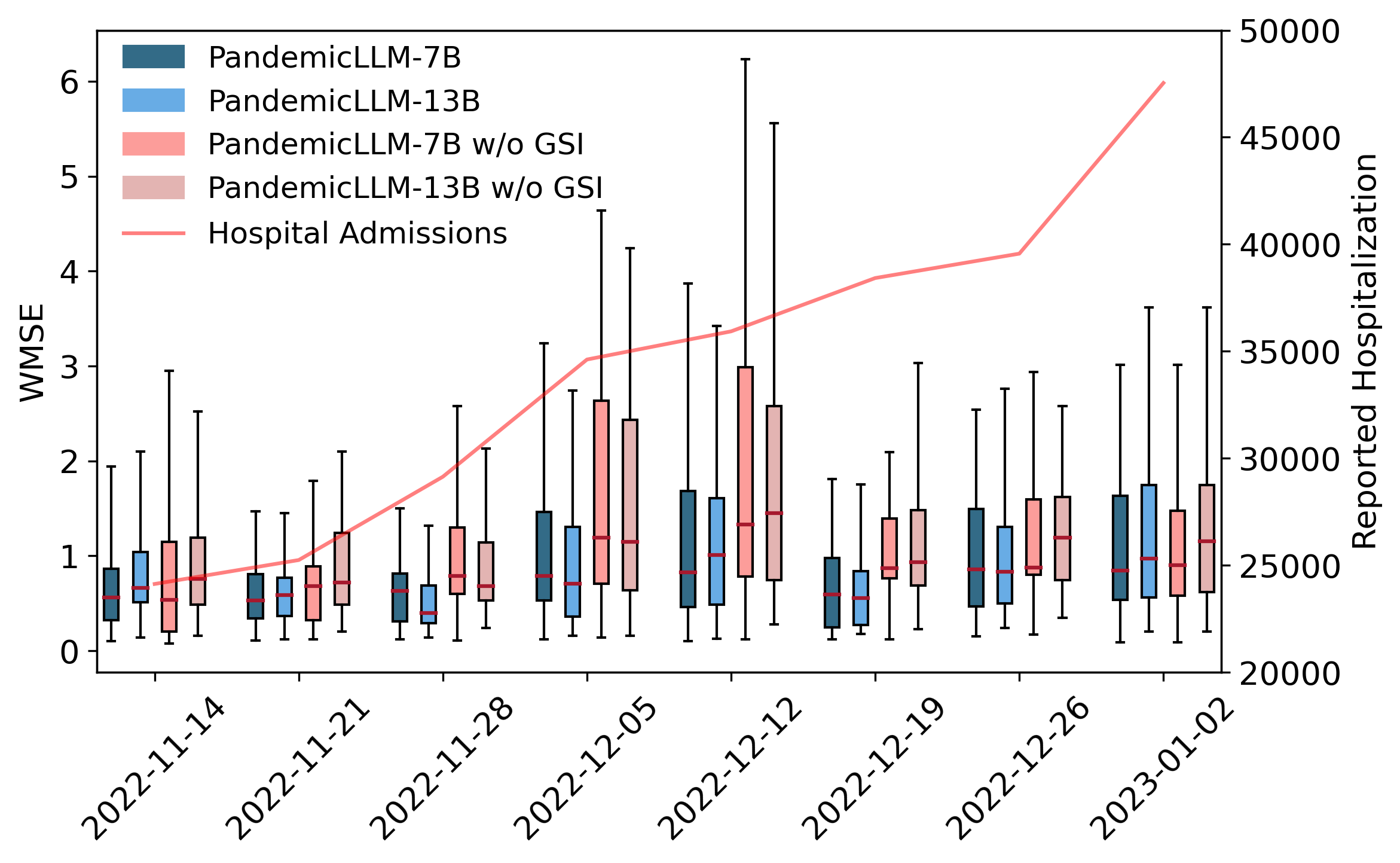}
  \caption{Model performance in terms of weighted MSE (WMSE).}
  \label{fig:variant_b}
\end{subfigure}
\begin{subfigure}{.5\textwidth}
  \centering
  \includegraphics[width=.95\linewidth,height=.6\textwidth]{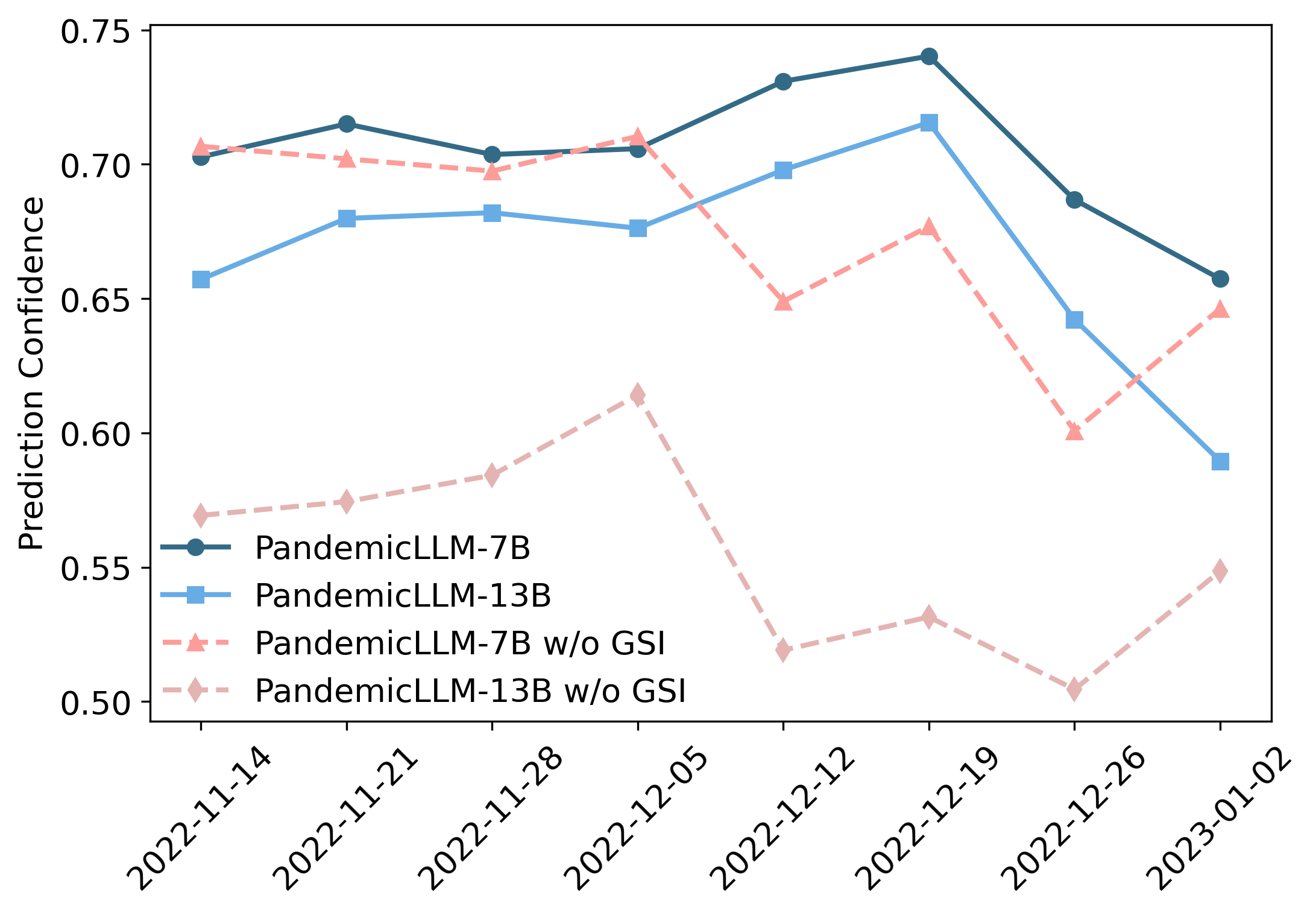}
  \caption{Model confidence (the probability of the predicted trend).}
  \label{fig:variant_c}
\end{subfigure}%
\caption{\textbf{A comparative analysis with and without the real-time genomic surveillance information.} \textbf{(a)} National estimates of weekly proportions of SARS-CoV-2 variants from September, 2022 to January, 2023. \textbf{(b)} Comparison of models' performance with and without real-time genomic surveillance information (w/o GSI). \textbf{(c)} Prediction confidence of \models across time. The dash lines represent the models without real-time genomic surveillance information.}
\label{fig:variant}
\end{figure}

Specifically, the hospitalizations surge during the testing period can be attributed to the rise of the SARS-COV-2 BQ.1 variant starting in October 2022, which became the predominant strain by December 2022, as depicted in Fig. \ref{fig:variant_a}. The initial authoritative report detailing the virological properties of the BQ.1 variant was released on October 27, 2022\cite{who_bq}. Subsequently, within the same week, we incorporated these specific characteristics of the BQ.1 variant (infectiousness, severity, and resistance to immunity), together with the latest variant proportion estimates from the CDC \cite{ma2023genomic}, into the \models (see Extended Data Fig. \ref{fig: example_genomic_prompt} for example genomic prompts).

The results indicate \models respond to the real-time genomic surveillance information. Fig. \ref{fig:variant_c} displays a comparison of four sets of distinct predictions: (1) Predictions generated by \model-7B, (2) Predictions generated by \model-7B without genomic surveillance information, (3) Predictions generated by \model-13B, and (4) Predictions generated by \model-13B without genomic surveillance information. Fig. \ref{fig:variant_b} highlights the models' prediction confidences change when variant data is included. The findings reveal that the integration of variant information enhances both the performance and confidence of the models, particularly in the case of \model-13B. Specifically, introducing variant information in \model-13B leads to an average increase in prediction confidence of 20.1\% and an average improvement of 28.2\% in WMSE. This enhancement in performance is particularly evident during the transition of the dominant variant from BA.5 to the BQ.1 lineage.

\section{Discussion}

\textbf{Reshaping pandemic forecasting by incorporating all disease-relevant data streams.} Traditional disease forecasting models heavily depend on structured numerical data, overlooking the wealth of information hidden within diverse disease-relevant sources. For instance, public health policies and real-time reports on emerging variants offer crucial information in textual formats that traditional models cannot access. To address this need, our proposed model unlocks the full potential of relevant information for pandemic forecasting, by reformulating it as a text reasoning problem and adapting LLMs. Through an AI-human cooperative prompt design, we integrate diverse disease-relevant data streams and formats – including public health policies (textual), epidemiological time series (sequential), genomic surveillance (textual and sequential), and local demographic and healthcare system data (numerical and categorical) – into well-structured prompts. This approach inclusively converts all information into text, allowing PandemicLLM to process and reason from data inaccessible to traditional frameworks. As demonstrated in section \ref{sec: model_comparison}, PandemicLLM achieves a significant performance increase of at least 20\% over existing models, highlighting the value of incorporating diverse data streams and formats within LLMs for pandemic forecasting. While our current model is a promising first step, we aim to expand its capabilities by including an even broader spectrum of disease-relevant data, such as wastewater-based epidemiology\cite{li2023wastewater} and human behavior data\cite{bedson2021review}, further enhancing its predictive accuracy and utility.


\textbf{Enhancing LLMs to master time series data and dependencies.} 
The proposed PandemicLLM framework is enhanced by the integrated sequential GRU encoder for hospitalization time series, leading to an accuracy improvement of 17\%- 24\% over the model without the encoder (see Extended Data table 3). This design allows for easy future extensions to accommodate other time series encoder architectures (such as LSTMs or attention mechanisms), potentially providing better suitability for different disease types and propagation patterns. The AI-human collaborative prompt design is aimed to enable automation in real-time, allowing human expertise to be strategically focused on selecting the most relevant and timely information from authoritative sources, such as policy updates and variant reports. We envision integrating our framework into ongoing and future efforts for real-time pandemic forecasting, leveraging the collaboration between AI efficiency and human judgment.

\textbf{Trustworthiness and robustness enhancement for better decision making.} For a model to effectively support public health decision-making, it needs to be reliable, trustworthy, and accurate\cite{thirunavukarasu2023large, singhal2023large}. Moreover, pandemic forecasts are inherently uncertain, and effectively communicating forecast uncertainty to decision-makers and the public is critical for transparency, yet remains challenging \cite{bertozzi2020challenges,nixon2022real}. In consideration of these needs, PandemicLLMs were fine-tuned to predict future pandemic trends by generating probabilities within defined categories, where the probability of the predicted category indicates the models’ confidence in their predictions. As revealed through section \ref{sec:Trustworthiness},~PandemicLLMs performance improves consistently as confidence increases. This observation is consistent with the capabilities of Flan-PaLM 540B\cite{singhal2023large}, an LLM with encoded clinical knowledge, further emphasizing the \models' proficiency in representing uncertainties with the COVID-19 related knowledge. Consequently, the confidence level can effectively function as an indicator of prediction reliability, offering model users a definitive guide to gauge the trustworthiness of the forecasts. Additionally, the confidence level can inform the sufficiency of available information. As evidenced in Fig. \ref{fig:variant_b}, including genomic surveillance information led to simultaneous improvements in both the models' performance and confidence. In light of our models' demonstrated strengths, they exhibit reliable performances and provide decision-makers with an incisive understanding of uncertainty. 

\textbf{Generalizing to other disease forecasts across diverse spatial-temporal scales.}
Section \ref{sec: variant} demonstrates PandemicLLM's zero-shot ability, allowing adaptation to emerging variants without retraining. Additionally, the 28.2\% performance improvement in WMSE with real-time genomic information highlights how PandemicLLMs leverage previously unseen variant information for reasoning. This attribute suggests the potential for generalizing our framework to other diseases with similar transmission mechanisms (such as flu and RSV) and adapting to emergency public health scenarios requiring rapid decision-making under limited data. The design of PandemicLLM allows for scalability and generalizability, not just in terms of the diseases it can predict but also in the granularity of its forecasts. Its success in capturing disease dynamics at the state level offers a promising foundation for extending its forecasts to more localized levels, such as counties or even hospitals. This potential adaptation would provide valuable insights into broader public health needs, enabling targeted forecasts that directly support local decision-making and interventions. As we look toward the future, the insights and methodologies developed through PandemicLLM offer a glimpse of what might be possible for the next generation of public health forecasting models. We envision future models building on this work, incorporating a wider variety of data, integrating AI and human expertise, and tackling an increasingly diverse array of public health challenges.

\section{Limitations}

A potentially limiting factor of the proposed model is due to the computational cost of employing LLMs. In efforts to address this concern, we empirically show that freezing the pre-trained LLaMA2 parameters achieves much better efficiency without sacrificing performance. However, \ours still demands a substantial amount of computational resources as the gradients of LLM parameters are involved in optimization of trainable RNNs and input token embeddings. This could be a limiting factor in scenarios where resources are scarce or in situations demanding rapid model deployment. 

Additionally, LLMs lack theoretical transparency. To address this issue we explicitly evaluate the empirical reliability and trustworthiness of \models. However, there remains a need to enhance LLMs’ interpretability. This is particularly important for public health related applications of LLMs, where elucidating the reasoning behind model predictions is pivotal for fostering user confidence and trust.

\section{Conclusion}


In this study, we introduced a novel real-time LLM-based framework for pandemic forecasting at the population level. The proposed PandemicLLMs extend the LLM architecture by integrating a temporal encoder tailored explicitly for processing epidemiological time series data. PandemicLLMs also incorporate unique disease-relevant data streams, such as textual public health policies and textual virological characteristics, which were previously inaccessible to existing forecasting models. Our findings demonstrate that PandemicLLMs outperform existing frameworks, offering robust and trustworthy predictions even for previously unseen scenarios, which can match the critical needs of public health policymakers. Through this work, we shed light on the potential of LLMs to improve strategies for pandemic response, envisioning a future where AI strengthens the resilience and efficiency of global health systems during public health crises.

\newpage 

\def\XX{\mathbb{X}}
\section{Methods}

\subsection{PandemicLLM datasets}
\label{sec:datasets}
The proposed \ours is fine-tuned using multiple disparate categories of data including spatial, epidemiological time series, genomic surveillance, and public health policy data. Our data covers all 50 states in the United States, ensuring a comprehensive nationwide scope for our study. All the spatial data are available at the state resolution, while all the time-varying data are available at the weekly resolution. In this section, we comprehensively discuss the sources of data and the pre-processing process.

\subsubsection{Epidemiological time series data}
\label{sec:temporal data}
\textbf{Hospitalizations.} Our study focuses on refining the targets used for COVID-19 forecasting models. The choice of hospitalization data over cases and deaths data stems from its capability to reflect the disease's spread and overall harmful impacts on healthcare systems, making it a more comprehensive measure.

Shifting from the conventional reliance on numerical targets, our study embraces categorical targets due to two main considerations: First, the unreliability of data reporting poses a significant challenge\cite{dong2022johns}. The inconsistencies in reported cases, deaths, and hospitalizations undermine the effectiveness of traditional disease forecasting models that rely heavily on these metrics. Second, the experience with the COVID-19 pandemic has demonstrated that the performance of numerical target predictions often falls short of expectations\cite{nixon2022evaluation, ioannidis2022forecasting}.

In light of these limitations, we advocate for the use of \textbf{hospitalization trend categories} (HTC) as our predictive targets instead of relying solely on numerical values, These categories, crafted from weekly COVID-19 hospitalization time series, offer a more robust indicator of both the disease spread and its impact on healthcare systems. We categorized the hospitalization numbers into five distinct trends as follows:

\begin{equation}
    HTC_1^{i, t} = \begin{cases}
\text{Substantial Increase} & \text{if $HT_1^{i, t} > 3$} \\
\text{Moderate Increase} & \text{if $3 > HT_1^{i, t} >1$} \\
\text{Stable} & \text{if $1 > HT_1i^{i, t} >-1$}\\
\text{Moderate Decrease} & \text{if $-1 > HT_1^{i, t} >-3$}\\
\text{Substantial Decrease} & \text{if $-3 > HT_1^{i, t}$}

    \end{cases}
\end{equation}

\begin{equation}
    HTC_3^{i, t} = \begin{cases}
\text{Substantial Increase} & \text{if $HT_3^{i, t} > 4.5$} \\
\text{Moderate Increase} & \text{if $4.5 > HT_3^{i, t} >1.5$} \\
\text{Stable} & \text{if $1.5 > HT_3^{i, t} >-1.5$}\\
\text{Moderate Decrease} & \text{if $-1.5 > HT_3^{i, t} >-4.5$}\\
\text{Substantial Decrease} & \text{if $-4.5 > HT_3^{i, t}$}

    \end{cases}
\end{equation}

Where $HTC_1^{i, t}$ and $HTC_3^{i, t}$ represents the 1 and 3-week later hospitalization trend category for state $i$ at week $t$. The 1 and 3-week hospitalization tends for state $i$ at week $t$ ($HT_1^{i, t}$ and $HT_3^{i, t}$) are defined as follow:

\begin{equation}
    HT_1^{i, t} = HR^{i,t} - \overline{HR}^{i,t-1}
\end{equation}

\begin{equation}
    HT_3^{i, t} =  HR^{i,t} - \overline{HR}^{i,t-3}
\end{equation}

Where $HR^{i,t}$ represents the hospitalization rate per one hundred thousand people for state $i$ at week $t$. $\overline{HR}^{i,t}$ indicates the smoothed 3-week average hospitalization rate per one hundred thousand people for state $i$ at week $t$. The decision to define hospitalization trends (HT) by contrasting current reported hospitalization values with lagged, smoothed data was to diminish the influence of data irregularities and reporting errors. This approach is intended to yield a robust and more representative target, thereby providing a more reliable reflection of the actual progression of the pandemic.

While the \target is utilized as the target, the $HT$ and the $HR$ are employed as input epidemiological time series data. The state-level COVID-19 hospitalizations data were collected from the U.S. Department of Health \& Human Service (HHS) \cite{HHS}. 

\textbf{Reported cases.} Our study utilized the state-level, daily reported COVID-19 case from the Johns Hopkins University Center for Systems Science and Engineering (CSSE) \cite{dong2020interactive}. The raw case data are aggregated at the weekly level as input data $C_i^t$, where $C_i^t$ represents the number of reported cases for state $i$ at week $t$. 

\textbf{Previous infections.} Numerous research findings have highlighted the protective role of prior infections in guarding against reinfection and severe outcomes of COVID-19 \cite{du2024association, altarawneh2022effects}. In order to incorporate the impact of immunity acquired from previous infections in mitigating severe disease upon subsequent infections, we created a variable that represents the total number of reported infections over a three-month period. The mathematical formulation of previous infection (PI) is defined as follows:

\begin{equation}
    PI_i^{t} = \frac{\sum_{j:t-16}^{t-4}C_i^j}{pop_i}
\end{equation}

Where $PI_i^{t}$ represents the cumulative reported infection rate for state $i$ from $16$ to 4 weeks prior to week $t$, $C_i^j$ is the number of reported cases for state $i$ at week $j$, and $pop_i$ is the population for state $i$. The numerator represents the aggregate of $C_i^j$ from the period spanning weeks $t-16$ to $t-4$ preceding week $t$.

\textbf{Vaccination.} The vaccine-induced immunity is widely regarded as a crucial approach to reducing the harmful impact of COVID-19. Our model incorporated state-level vaccination data from the US CDC Vaccine Tracker \cite{cdc_vaccination}, leveraging this information to enhance our analysis. We included state-level cumulative vaccination rates normalized by population. The vaccination data covers three types of time series: the partial vaccination rate, the completed primary series rate, and the booster vaccination rate. Our approach offers a detailed view of the vaccination landscape across different states by including these distinct metrics.

\subsubsection{Spatial data}

\textbf{Demographics.} Data on the state population, including age group and race-related data, was sourced from the American Community Survey by the US Census Bureau \cite{Bureau_2023}. Recognizing that COVID-19 affects different racial and age groups unevenly, as established by existing studies \cite{dowd2020demographic, bollyky2023assessing}, we specifically focused on the population aged over 65 and vulnerable racial groups. The demographic information utilized in our research is derived from the 2022 National Census Survey.

\textbf{Healthcare systems.} In our study, we incorporated healthcare system scores from the Commonwealth Fund \cite{radley2022commonwealth}, an annually updated and comprehensive dataset that evaluates the performance of healthcare systems in every U.S. state. This score on State Health System Performance offers a unique perspective on how state healthcare systems have coped with and managed the challenges posed by the COVID-19 pandemic. For each state, this resource provides not only an overall healthcare system performance ranking but also offers detailed rankings in specific categories such as COVID-19 response, Access and affordability, Prevention and treatment, Avoidable Hospital Use and Cost, Healthy Lives, Income Disparity, and Racial and Ethnic Equity. 


\textbf{2020 presidential election results.} Prior research has identified a notable association between political affiliation and COVID-19 health outcomes at the state level \cite{bollyky2023assessing}. In light of this, we collected data for the 2020 U.S. Presidential election results from the Federal Elections Commission \cite{election}. Based on the state-level voting results, we labeled each state as either Democrat or Republican.

\subsubsection{Public health policy data.} The impact of the COVID-19 epidemic is influenced by the stringency and timing of government-implemented policies\cite{haug2020ranking}. These policies include measures such as closing schools, canceling public events, and protecting the elderly vulnerable populations. We collected policy data from the Oxford COVID-19 Government Response Tracker \cite{hale2021global}. However, in contrast to the majority of studies that utilize the generated policy index, we integrate the textual descriptions of selected policies as input for our prompt design, providing a unique approach to analyzing policy impact. In order to effectively assess the impact of policies on the spread of the disease and the strain on healthcare systems, we specifically chose policies from two categories: 'C' for containment and closure policies and 'H' for health system policies. The complete list of policies included in our study from these categories is provided in Extended Data Table \ref{tab:policies}.

Our prompt utilized the summary of the stringency levels for each policy as input. These summaries are derived directly from the original descriptions provided by the data source. Due to restrictions in token size, we have limited our input to the specified policy categories despite the potential impact of other policies on the pandemic's dynamics.

\subsubsection{Genomic surveillance data.}
In the constantly changing COVID-19 pandemic, the emergence of new variants plays a pivotal role in shaping pandemic trends and potential new waves. Genomic surveillance data is indispensable in this context, as it enables the timely detection and tracking of these variants, providing essential insights into their characteristics and spread \cite{stockdale2022potential}. Existing forecasting models that attempt to incorporate genomic surveillance data frequently encounter limitations due to reporting delays and the challenge of accurately encoding the virological characteristics of different variants \cite{du2023incorporating, rashed2022covid}. \ours offer a novel opportunity to include this data more effectively. Through zero-shot inference, LLMs can process and interpret previously \textit{unseen} variant data without training, providing a fresh and timely perspective in dealing with emergence of new variants. In our study, we utilized two primary types of genomic surveillance data to forecast the COVID-19 \target:

\begin{itemize}
    \item \textbf{Virological characteristics:} Our study integrated official updates from authoritative organizations, including the World Health Organization (WHO) \cite{WHO}, the Centers for Disease Control and Prevention (CDC) \cite{CDC}, and the European Centre for Disease Prevention and Control (ECDC) \cite{ECDC}. These updates, presented in text format, offer information on the virological characteristics of newly emerging variants. We then summarized the information from three aspects: 1) infectiousness, 2) severity, and 3) resistance to immunity. More detailed data sources are documented in the Supplementary Information section 1. 
    
    \item \textbf{Weighted estimates of variant proportions:} We also included the weighted estimates of variant proportions from the CDC \cite{ma2023genomic} in our analysis. Weighted estimates refer to the proportions of circulating variants derived from empirical genomic sequencing data that has been observed and recorded. This data is vital for assessing the prevalence and distribution of these variants, offering an overview of the pandemic's status and aiding in more accurate forecasting.
\end{itemize}

\subsection{Preliminaries and notations}
\label{sec: preliminary}

\textbf{Large language models.} In recent years, Large Language Models (LLMs)\cite{GPT3,ChatGPT,PaLM2,GPT4} have catalyzed a pivotal transformation in the field of natural language processing. These models are transformer decoders\cite{Transformer} autoregressively pre-trained on extensive text corpus such as Github, Wikipedia, CommonCrawl, and BooksCorpus. Consequently, LLMs enjoy a broad knowledge base and demonstrate proficiency across a spectrum of tasks, ranging from conversational agents~\cite{ChatGPT} to complex reasoning~\cite{ComplexCoT} and decision-making applications~\cite{ReAct, CodeAsPolicies}. Their performance often rivals, and occasionally surpasses, that of humans in these domains~\cite{GPT4}. Despite the remarkable performance of LLMs, the state-of-the-art LLMs are often closed-source LLMs, e.g. ChatGPT~\cite{ChatGPT}, and GPT-4~\cite{GPT4}, hampering the development and application of LLMs for the research community. To bridge this gap, LLaMA~\cite{LLaMA} and LLaMA2~\cite{Llama2} are proposed. They are open-source LLMs trained on publicly available datasets and achieved comparable or better performance than GPT-3. In this study, we used LLaMA2 as the backbone for its robust performance in language understanding. 

\textbf{Autoregressive language modeling.} Most existing LLMs are trained in an autoregressive manner, where the model predicts the next token in a text sequence based on its predecessors. This process begins with tokenization, e.g. Byte Pair Encoding (BPE)~\cite{BPE}, where a tokenizer segments a string into discrete tokens. Then, the tokenized sequence for data sample $i$, denoted as $T_i$ is reconstructed autoregressively: 
\begin{equation}
   p_{\theta}(T_i)=\prod_{j=1}^{|T_i|}{p_{\theta}(t_j^{(i)} \vert t_{1}^{(i)}, \cdots , t_{j-1}^{(i)}}),
\label{eq: autoregressive}
\end{equation}

where $p_{\theta}$ is the LLM model, $t_j^{(i)}$ denotes the $j$-th token in $T_i$. By maximizing the likelihood $ p_{\theta}(T)=\prod_{i=1}^{N}{p_{\theta}(T_i)}$ for all data samples $T_i \in T$, the LLM's parameters are learned.

\textbf{Challenges of LLMs in encoding disease-related data.} 
Despite the remarkable abilities of LLMs, pandemic forecasting integrates various data sources, including demographic profiles, public health policy, and epidemiological time series. These diverse sources constitute a composite, multi-modal input for the pandemic prediction model. However, conventional LLMs primarily handle discrete text inputs, which can not directly leverage these multi-modality data. Besides, for continuous sequential data, tokenizers like Byte Pair Encoding (BPE) segment continuous sequences into discrete tokens, breaking numbers into awkward chunks that make learning basic numerical operations challenging~\cite{LLM_time}. To leverage the cross-domain knowledge of LLMs while preserving sequential information effectively, it is essential to innovate how multi-modal pandemic data is assimilated as input for LLMs.


\textbf{Notations.} Before we formally start to introduce the \ours we establish the following notations. 

\begin{itemize}
    \item $T_i$ denotes the prompt text of the $i$-th item, including task description, spatial, epidemiological time series, genomic surveillance, and public health policy information. An example can be found in Figure~\ref{fig: fig_2}.
    \item $X_i$ denotes the time series data of the $i$-th item in the dataset, e.g. hospitalization time-series data.
    \item $y_i$ denotes the token for Hospitalization Trend Category (HTC) of the $i$-th item in the dataset, representing one of the following trends \textit{\{<Substantial Decrease>, <Moderate Decrease>, <Stable>, 
<Moderate Increase>, <Substantial Increase>\}}.
    \item $f_\theta$ denotes an LLM with parameters $\theta$. $\theta_{in}$ and $\theta_{out}$ denote the parameters for the input and output layer.
    \item $g_{\phi}$ denotes the RNN encoder with parameters $\phi$.
    \item $\bm{H}_{i}$ denotes the embedding matrix of text tokens $T_{i}$ that serves as the input for the transformer layers.
    \item $\boldsymbol{z}_{i}$ denotes the embedding vector of time series data $X_i$. 
\end{itemize}

\subsection{Methodology}



\subsubsection{Overview of \ours}
We propose to formalize the pandemic prediction problem as a multi-modality ordinal classification problem: For a sample $i$ (e.g., New York state at a given week), given the prompt with multi-modality text information, denoted as $T_i$, and the sequential data $X_i$, the goal is to predict the Hospitalization Trend Category $y_i$. Specifically, as illustrated in Extended Data Fig.~\ref{fig: model_framework}, \ours models $p(y_i|T_i, X_i)$ using two models: an RNN-based sequential encoder $g_\phi$ that projects sequential data to the representations in the text space; A transformer-based LLM $f_\theta$ that forecasts the categorical targets distribution using the encoded text representations containing multi-modality information and the encoded time series information. In this way, \ours seamlessly integrates text reasoning and numerical time series learning using one unified framework.

\subsubsection{Multi-modality data textualization}
\label{prompt_design}

\begin{figure}[h!]
    \centering
    \includegraphics[width=0.8\linewidth]{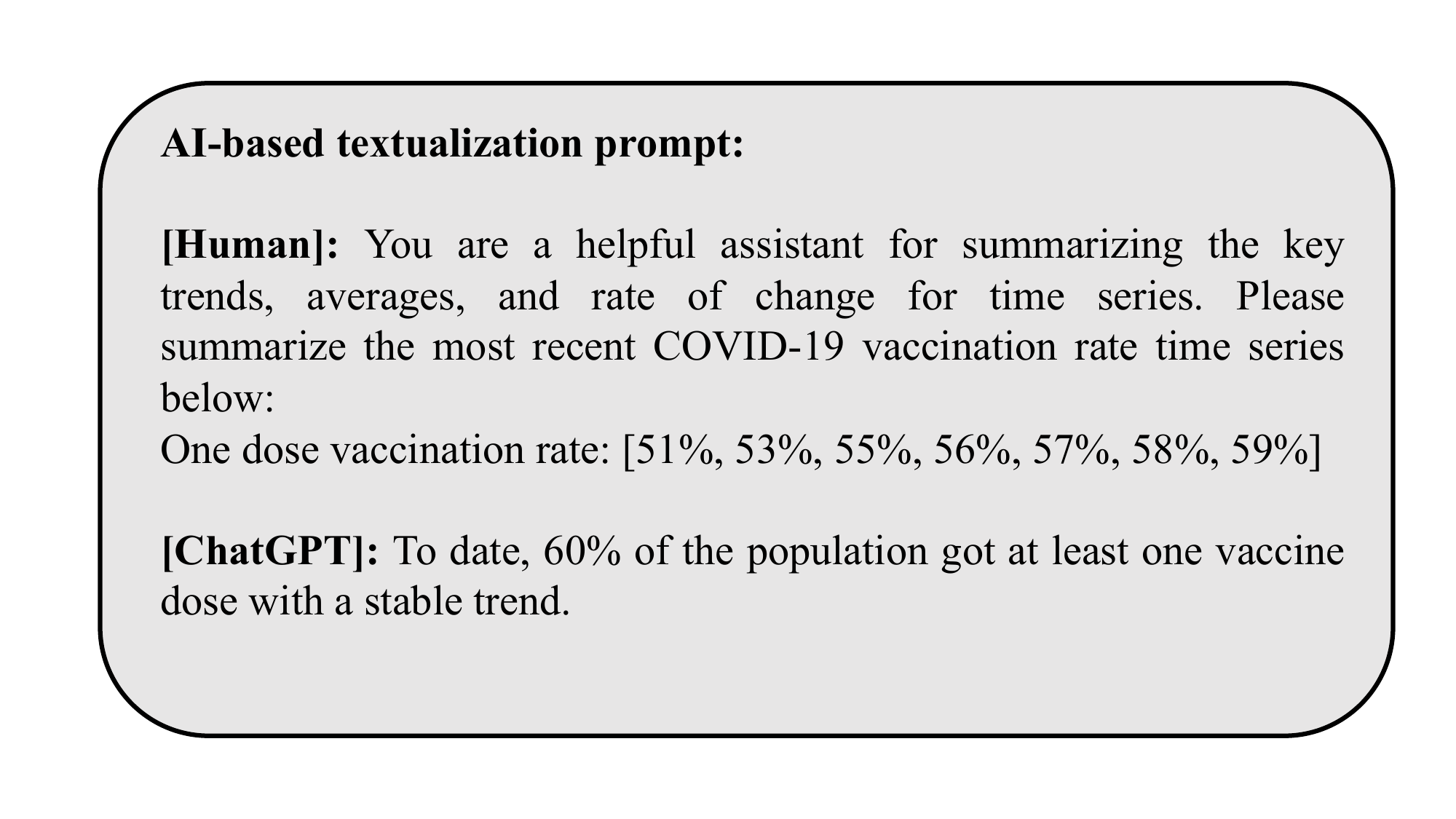}
    \caption{AI-based textualization: \ours utilizes an LLM, e.g. ChatGPT, to translate the numerical sequence data into detailed textual narratives. The figure takes COVID-19 vaccination time series data as an example.}
    \label{fig: AI_textualization}
\end{figure}

\begin{figure}[h!]
    \centering
    \includegraphics[width=0.7\linewidth]{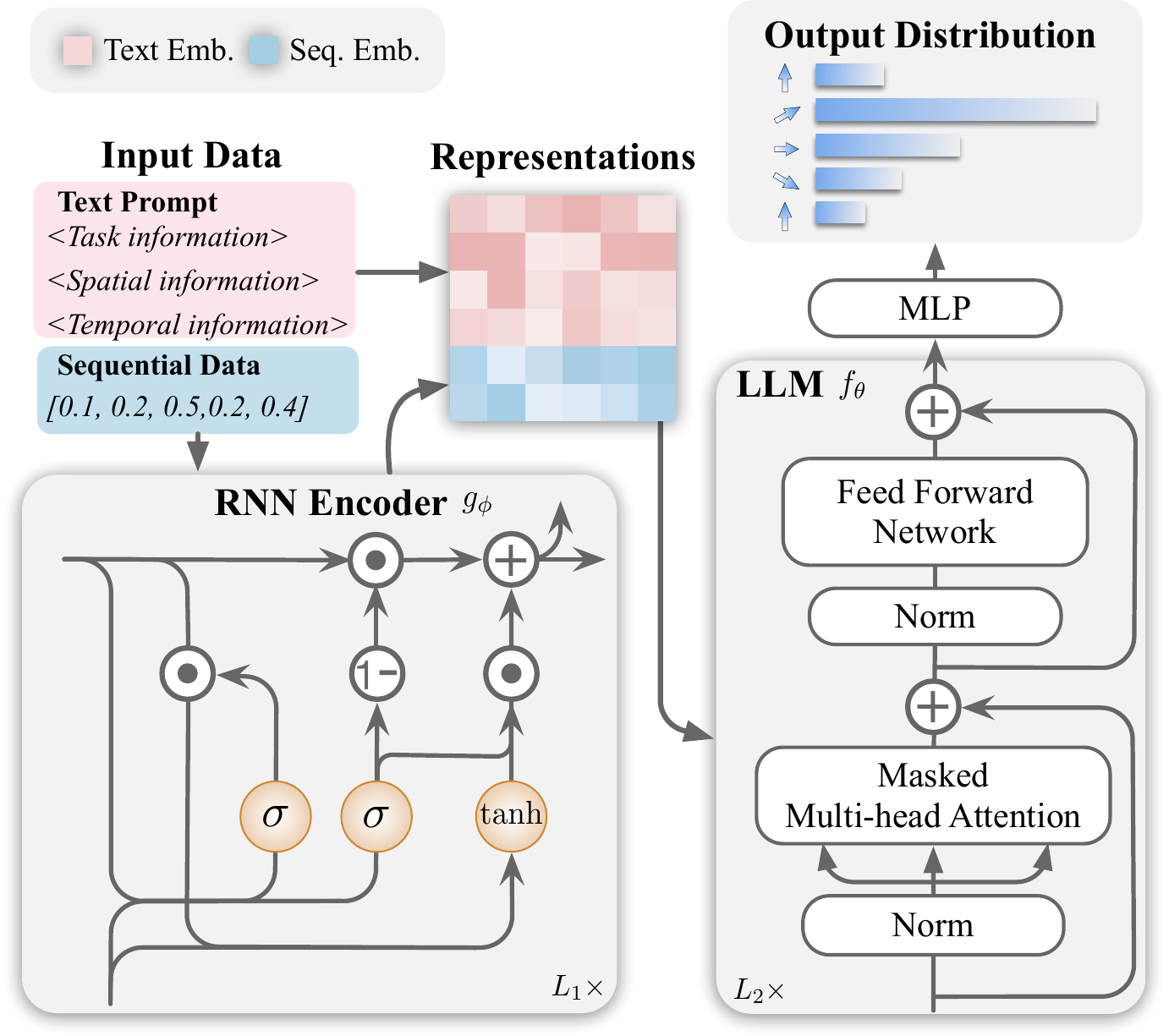}
    \caption{The \ours Model: The input for \ours contains two modalities : (1) Textualized multimodal prompt consisting of task information, spatial information and temporal information (2) Sequential data such as hospitalization and recent reported cases. \ours looks up the text representations and encodes the sequential data as sequential representations. The combined representation is further fed to an LLM to forecast the distribution of the pandemic trend. }
    \label{fig: model_framework}
\end{figure}

\begin{figure}[h!]
\centering
    \includegraphics[width=0.8\textwidth]{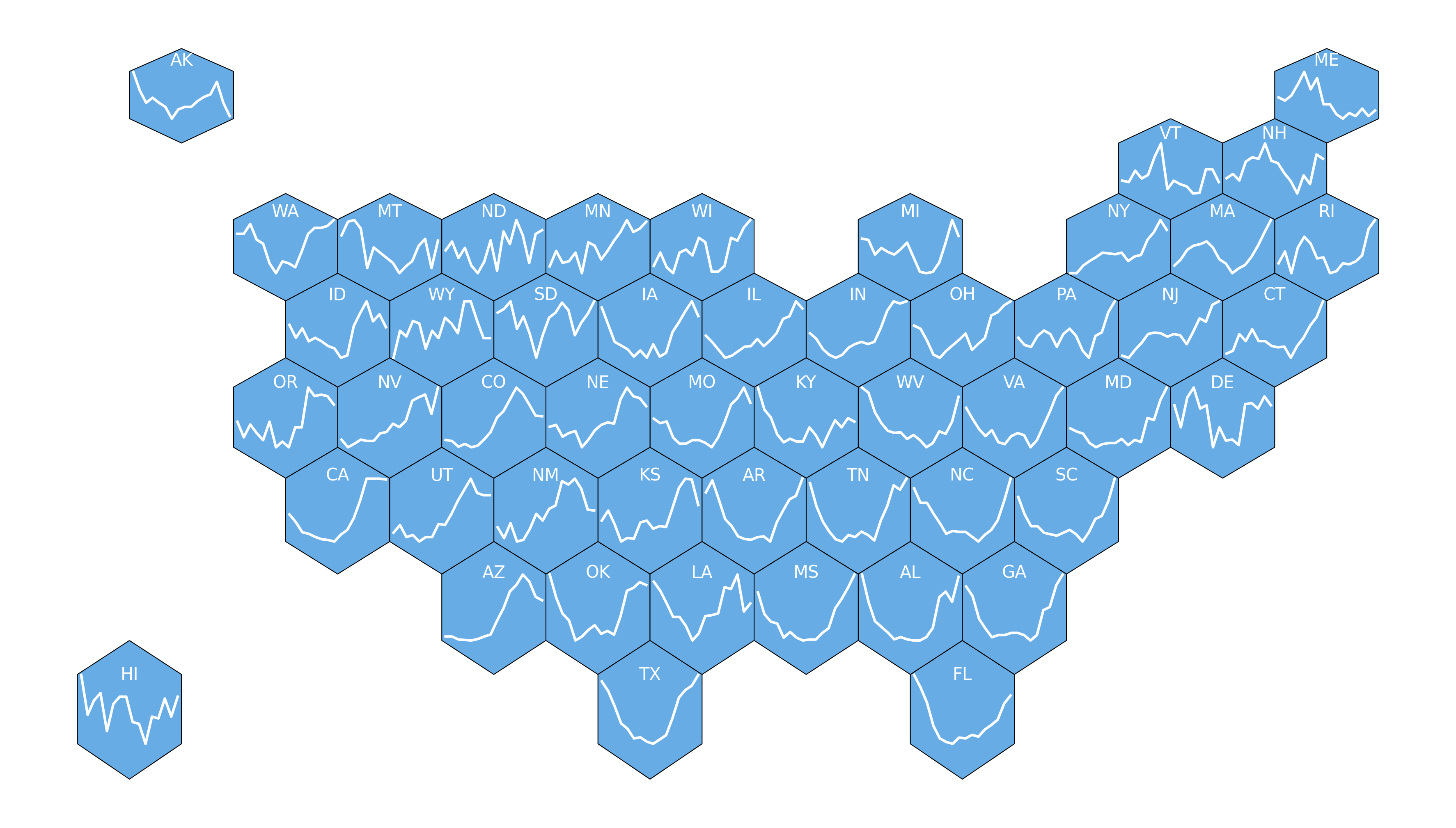}
    \caption{The COVID-19 hospitalization trends from September 2022 to January 2023.}
    \label{fig: hosp_trend}
\end{figure}

\begin{figure}[h!]
    \centering
    \includegraphics[width=0.9\linewidth]{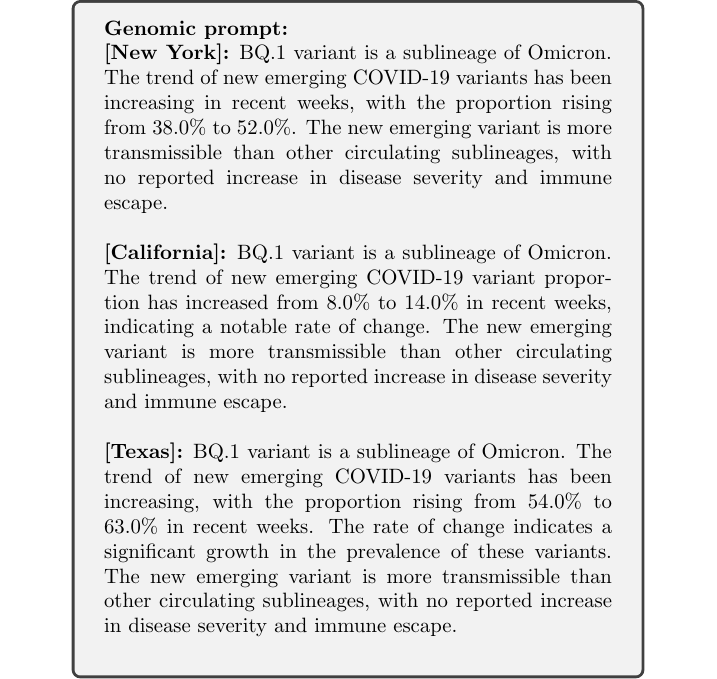}
    \caption{Example genomic prompts. \models provide timely responses to genomic surveillance information by a prompt containing virological characteristics and recent trends of variants' proportion estimates. The figure illustrates three representative states during the period of the BQ.1 variant emergence.}
    \label{fig: example_genomic_prompt}
\end{figure}

\begin{figure}[h!]
    \centering
    \includegraphics[width=0.9\linewidth,height=20cm]{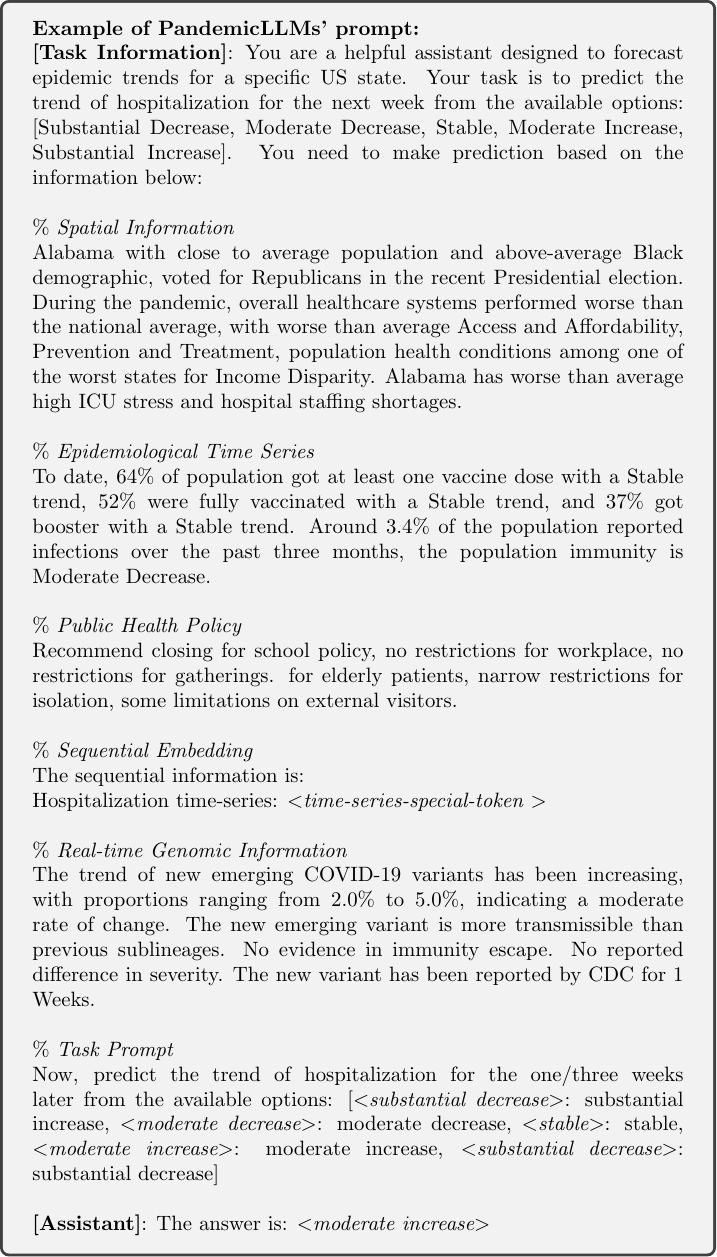}
    \caption{\ours organizes multimodality information for the pandemic in a text prompt, comprehensively covering the spatial, temporal, sequential, genomic, and relevant tasks instructions for LLMs to perform pandemic reasoning.}
    \label{fig: example_prompt}
\end{figure}

\ours utilizes an LLM to address the challenge of processing multi-modal disease-relevent data. For each item in the dataset, we construct a composite prompt to effectively parse and integrate the multi-modality data. An example is shown in Extended Data Fig. \ref{fig: example_prompt}:

\begin{itemize}
    \item \textbf{Spatial data textualization:} In spatial data textualization, we processed numerical state demographic data and healthcare system statistics. Those data for all 50 U.S. states were numerically ranked and subsequently transformed into categorical descriptions reflecting their relative positions. States ranking in the top five were labeled "One of the best." Those between $6^{th}$ and $20^{th}$ were categorized as "Higher than the national average," while rankings from $21^{st}$ to $30^{th}$ were described as "Close to the national average." States falling between $31^{st}$ and $45^{th}$ were labeled "Lower than the national average," and those in the bottom five were described as "One of the lowest."  Regarding presidential election outcomes, each state was characterized based on its voting percentage, being labeled as either predominantly voting for Democrats or Republicans (see Supplementary Information section 2).
    \item \textbf{Epidemiological time series data textualization:} 
In the phase of textualizing epidemiological time series data, an AI model (ChatGPT-3.5~\cite{ChatGPT}) is employed to convert sequential data into narrative summaries. This method leverages the sophisticated capabilities of AI to transform numerical sequences into detailed textual narratives, thereby reducing the need for laborious manual annotation. For instance, the AI-facilitated textualization process for the weekly vaccination rate time series is illustrated in Extended Data Figure \ref{fig: AI_textualization}.
    
    \item \textbf{Epidemiological time series representation learning:} As time series plays a critical role in pandemic forecasting, we leverage an additional RNN encoder to effectively distill the useful information into LLM's input space. Specifically, the prompt for time-series data is initialized by a special token \textit{<time-series-special-token>} whose embedding $\boldsymbol{z}_i$ is encoded by an RNN ecnoder $g_{\phi}$:
\begin{equation}
    \boldsymbol{z}_{i}=g_{\phi}(X_i) \in \mathbb{R}^{d},
\end{equation}
    where $X_i$ denotes the time-series data, e.g. hospitalization time-series, for data sample $i$. As the Extended Data Table \ref{tab:ablation} showed, the RNN encoder can significantly improve the model's performance. And we choose GRU as the implementation of $g_{\phi}$ in our experiments.
    \item \textbf{Policy textualization:} Our study incorporated six policy categories: School and Workplace Closures, Public Events, Gathering Restrictions, Facial Coverings, and Elderly Protection. We designed the prompt's weekly policy stringency level descriptions, highlighting policy shifts (see Supplementary Information section 2). For instance, if New York shifted its school closure policy from "recommended closures" to "No measures" while maintaining "No restriction" on gatherings, the description would be: "There have been changes in school policy moving from recommended closures to no restrictions while gathering policy remains unrestricted."
    \item \textbf{Genomic surveillance data textualization:} The genomic surveillance data consisted of two main categories: reports about new variants from authoritative sources and sequential data derived from the CDC's weighted estimates of variant proportions Key information, includes the variant's relative transmissibility, severity and the impact on immunity, was incorporated directly into the prompt design. The weighted variant proportion estimates were processed using the same method outlined in the temporal data textualization section (for further details, refer to Supplementary Information Sections 1 and 2).


\end{itemize}

Now we describe how we organize the multi-modality data into the input space of an LLM: we first provide the essential task information at the beginning of the prompt, then we use expert knowledge combined with AI assistance to describe the spatial and temporal information for pandemic forecasting. Then the sequential data is followed, which will be further encoded by the RNN encoder. The input for LLM transformer-decoder is a mixture of text information and encoded sequential information.

Specifically, for data sample $i$, the tokenized text $T_i$, containing both spatial and temporal information, is encoded into the text embeddings $\bm{H}_i$:

\begin{equation}
    \bm{H}_i = f_{\theta_{in}}(T_i) \in \mathbb{R}^{|T_i| \times d},
\end{equation}
where $f_{\theta_{in}}$ is the input embedding layer for LLM, $|T_i|$ denotes the number of tokens for $T_i$, $d$ is the embedding dimension of the LLM. 
Then, the encoded information $\boldsymbol{z}_{i}$ is used to replace the embedding of the time-series special token, i.e. \textit{<time-series-special-token>}, in $\bm{H}_i$ to generate the final representation for the transformer's input $\bm{H}_i'$:
\begin{equation}
\begin{aligned}
    \bm{H}_i'[j,:] =\left\{
    \begin{array}{ll}
        \bm{H}_i[j,:] & \quad \text{if } j \neq s_i, \\
        \boldsymbol{z}_{i} & \quad \text{if } j = s_i,
    \end{array}
    \right.
\end{aligned}
\end{equation}
where $s_i$ denotes the index for time series special token. In this way, \ours seamlessly fuses the sequential information and textual information into the input space of an LLM, enabling it to perform reasoning with both types of information.

\subsubsection{LLM for pandemic prediction}
\label{fine-tune}
With $\bm{H}_i'$ encoding both textual and sequential information, \ours leverages an LLM (i.e. transformer decoder) to perform pandemic prediction as a text generation problem. As introduced in Section~\ref{sec: preliminary}, an autoregressive LLM generates one token at a time conditioned on the previously generated tokens. To make a prediction, we extend the original vocabulary of LLM with  ``class tokens'': \textit{\{<Substantial Decrease>, <Moderate Decrease>, <Stable>, 
<Moderate Increase>, <Substantial Increase>\}} that represent the trend for future hospitalization. Then, we use the predicted distribution of ``class tokens'' after the output prompt ``The answer is'' as the target distribution for the pandemic forecasting. In this way, \ours formulates pandemic forecasting as a text reasoning problem using an LLM.

The model is optimized by the autoregressive loss defined in Eq.~\ref{eq: autoregressive}, maximizing the likelihood of the ground truth text. For better scalability and efficiency we freeze the transformer parameters for LLMs~\cite{LLaVA} and only train the vocabulary embeddings $f_{\theta_{in}}$ and output prediction layer $f_{\theta_{out}}$, as well as the GRU encoders $g_\phi$. 


\subsubsection{Evaluation}

\label{Evaluation}
We comprehensively evaluate our model using five error metrics: a) Accuracy, b) Mean Square Error (MSE), c) Weighted Mean Square error (WMSE), d) Brier Score, and e) Rank Probability Score (RPS). 

Accuracy offers a direct method to assess the performance of a model's classification, defined by the following equation:

\begin{equation}
\text{Accuracy} = \frac{\sum_{i=1}^{N} [y_i = \hat{y}_i]}{N},
\end{equation}

where \(y_i\) represents the actual class, \(\hat{y}_i\) denotes the predicted class, and \(N\) signifies the total number of samples. However, despite being widely used in traditional classification problems, equally treats all errors and neglects the inherent ordering between the HTC classes. For example, if the ground truth trend is ``Substantial Increase'', accuracy will treat the prediction error of ``Moderate Increase'' and ``Substantial Decrease'' equally.

To model the order information of HTC classes, we map the classes: \{Substantial Decrease, Moderate Decreasing, Stable, Moderate Increasing, Substantial Increase\} into numeric scale \{1, 2, 3, 4, 5\}, and use the mapped value to compute numerical error metrics. To start with, the mean squared error (MSE) is a promising way to evaluate the ordinal classification~\cite{OrdClsEval}, which is defined as:

\begin{equation}
    \text{MSE} = \frac{1}{N} \sum_{i=1}^{N} (\tilde{y}_i - \bar{y}_i)^2,
\label{eq:MSE}
\end{equation}

where $\tilde{y}_i$ and $\bar{y}_i$ denote the numerical values of the ground truth class and the predicted class. To further evaluate the predicted distribution, we use weighted MSE (WMSE), which introduces a probability weighting into the MSE, defined as:

\begin{equation}
    \text{WMSE} = \frac{1}{N} \sum_{i=1}^{N}\sum_{k=1}^K P(\bar{y}_i=k)(k - \hat{y}_i)^2,
    \label{eq:WMSE}
\end{equation}

where $K$ denotes the number of classes ($K=5$ for HTC prediction), $P(\bar{y}_i=k)$ denotes the predicted probability for the $k$-th class for the $i$-th data sample.


The Brier Score is used to measure the accuracy of probabilistic predictions. It is calculated as the mean squared difference between the predicted probability assigned to the possible outcomes and the actual outcome:
\begin{equation}
    \text{Brier Score} = \frac{1}{N} \sum_{i=1}^{N} \sum_{k=1}^{K} (P(\bar{y}_i=k) - o_k^{(i)})^2
    \label{eq:Brier}
\end{equation}
where $P(\bar{y}_i=k)$ is the forecast probability for the $i$-th item in the $k$-th class, and $o_k^{(i)}$ is the one-hot encoded actual outcome, with a value of 1 for the ground truth class and 0 for all other classes. 

We also utilize the Rank Probability Score (RPS) \cite{leung2021real}, which is particularly useful for evaluating the accuracy of the entire predicted probability distribution across categorical outcomes. The RPS is defined as: 

\begin{equation}
    \text{RPS} = \frac{1}{N} \sum_{i=1}^{N} \sum_{k=1}^{K} (q(\bar{y}_i=k) - \bar{q}_k^{(i)})^2 ,
    \label{eq:rps}
\end{equation}

where $q_k^{(i)}$ and $\bar{q}_k^{(i)}$ are the ground truth and predicted cumulative probability distributions for the $i$-th case in the $k$-th class. The ground truth cumulative probability is defined as a step function that increases from 0 to 1 at the class of the actual observed outcome. The predicted cumulative probability distribution is defined as the sum of the predicted probabilities for all classes up to and including the $k$-th class.



%









\newpage 

\section{Data availability}

All data utilized in this study derive from publicly accessible sources. Details of each raw data source and data processing are described in the Method Section. The processed data are available at \url{https://github.com/miemieyanga/PandemicLLM}. 
\section{Code availability}

Code is publicly accessible at \url{https://github.com/miemieyanga/PandemicLLM}. 
\bibliography{Method}

\section{Acknowledgment}

This research is supported by NSF Award ID 2229996, NSF Award ID 2112562 and ARO W911NF-23-2-0224. 
\section{Author Contributions}
H.D., J.Z. and H.F.Y conceptualized and designed the study. H.D. and S.X. collected data. H.D. processed the data and designed prompts. J.Z. and Y.Z. performed experiments. S.X. run the baseline models. H.D., J.Z., Y.Z., S.X. and H.F.Y prepared the figures. H.D., J.Z., Y.Z., and H.F.Y analyzed results. H.D., J.Z., Y.Z. and H.F.Y wrote the initial draft. L.M.G., Y.C., X.L. and H.F.Y provided guidance and feedback for the study.  L.M.G, X.L and H.F.Y revised the manuscript. L.M.G. and Y.C. acquired the funding. Y.C. provided computational resources. All authors prepared the final version of the manuscript. 

\section{Competing interests}
The authors declare no competing interests.

\newpage \section{Extended Data}

\begin{table}[H]
    \begin{center}
    \caption{Summary of policy types and stringency levels. This study includes a range of policies, particularly emphasizing containment, closure, and health system policies. A complete list of included policies is presented in the following table.  }
    \label{tab:policies}
    \small
    \setlength{\tabcolsep}{4pt} 
    \renewcommand{\arraystretch}{1.1} 
    \resizebox{0.8\textwidth}{!}{
    \begin{tabular}{c|c|c}
    \toprule[1.5pt]
    \textbf{ID} & \textbf{Policy type} & \textbf{Summary of stringency level}\\
    \hline
    \multirow{4}{*}{\makecell[c]{C1}} & \multirow{4}{*}{School closing} & No measures. \\
    & & Recommend closing.\\
    & & Require closing at some levels.\\
    & & Require closing at all levels.\\
    \hline
    \multirow{4}{*}{\makecell[c]{C2}} & \multirow{4}{*}{Workplace closing} & No measures. \\
    & &Recommend closing.\\
    & & Require closing for some categories.\\
    & & Require closing for all but essential workplaces.\\
    \hline
    \multirow{3}{*}{\makecell[c]{C3}} & \multirow{4}{*}{Public events} & No measures. \\
    & & Recommend canceling. \\
    & & Required canceling. \\
    \hline
    \multirow{5}{*}{\makecell[c]{C4}} & \multirow{5}{*}{Restrictions on gatherings} & No measures. \\
    & & Restrictions on very large gatherings.  \\
    & & Restrictions on gatherings between 101-1000 people \\
    & & Restrictions on gatherings between 11-100 people. \\
    & & Restrictions on gatherings of 10 people or less. \\
    \hline
    \multirow{4}{*}{\makecell[c]{H8}} & \multirow{5}{*}{Protecting elderly people} & No Measures. \\
    & & Recommended isolation and visitor restriction.\\
    & & Narrow restrictions for isolation and some limitations on external visitors.\\
    & & Extensive restrictions for isolation and all non-essential external visitors prohibited.\\
    
    \bottomrule[1.5pt]

    \end{tabular}
}
    \end{center}
\end{table}

\begin{table}[H]
    \begin{center}
    \caption{Ablation study for the RNN encoder of \model-7B focusing on 1-week prediction results. Here, $\uparrow$/$\downarrow$ signifies that a larger/smaller metric indicates superior performance. Boldface highlights the top performances, while parentheses detail the performance differences relative to the GRU-based model.}

    \label{tab:ablation}
    \small
    \setlength{\tabcolsep}{4pt} 
    \renewcommand{\arraystretch}{1.1} 
    \resizebox{0.9\textwidth}{!}{
    \begin{tabular}{c|c|c|c|c|c}
    \toprule[1.5pt]
 \multirow{2}{*}{\textbf{RNN Encoder Type}} & \multicolumn{5}{c}{\textbf{Evaluation Metric}}\\
     \cline{2-6}
      & \textbf{Accuracy $\uparrow$} & \textbf{MSE $\downarrow$} & \textbf{WMSE $\downarrow$}  & \textbf{Brier Score $\downarrow$} & \textbf{RPS$\downarrow$}\\\hline

\textbf{GRU} & \textbf{0.554} & \textbf{0.593} & \textbf{0.668} & \textbf{0.668} & \textbf{0.098} \\
vanilla RNN & 0.549(-0.005) & 0.631(+0.038) & 0.678(+0.010) & 0.692(+0.024) & 0.102(+0.004) \\
LSTM & 0.522(-0.032) & 0.650(+0.057) & 0.756(+0.088) & 0.681(+0.013) & 0.104(+0.006) \\
w/o RNN encoder & 0.446(-1.080) & 0.925(+0.332) & 2.301(+1.633) & 0.752(+0.084) & 0.151(+0.053) \\
      
    \bottomrule[1.5pt]

    \end{tabular}
    }
    \end{center}
\end{table}

\end{document}